\documentclass[final,5p,times,twocolumn,authoryear]{elsarticle}
\usepackage{amssymb}
\usepackage{graphicx}
\usepackage{booktabs}
\usepackage{multirow}
\usepackage{amsmath}
\usepackage{microtype}

\usepackage{mathtools}
\usepackage{bbm}
\usepackage{float}
\usepackage{rotating} 
\usepackage[ruled,vlined]{algorithm2e}

\usepackage[table,xcdraw]{xcolor}
\usepackage{colortbl} 
\usepackage{amsmath}
\usepackage{orcidlink}
\usepackage{subcaption}

\journal{}

\begin{document}

\begin{frontmatter}

\title{Improving Adversarial Robustness for 3D Point Cloud Recognition at Test-Time through Purified Self-Training}

\author[1]{Jinpeng Lin}
\author[2]{Xulei Yang}
\author[1]{Tianrui Li}
\author[2]{Xun Xu\corref{cor1}}

\address[1]{School of Computing and Artificial Intelligence, Southwest Jiaotong University, China}
\address[2]{Institute for Infocomm Research (I2R), A*STAR, Singapore}

\cortext[cor1]{Correspondence to: Xun~Xu e-mail: alex.xun.xu@gmail.com .}

\begin{abstract}
Recognizing 3D point cloud plays a pivotal role in many real-world applications. However, deploying 3D point cloud deep learning model is vulnerable to adversarial attacks. Despite many efforts into developing robust model by adversarial training, they may become less effective against emerging attacks. This limitation motivates the development of adversarial purification which employs generative model to mitigate the impact of adversarial attacks. In this work, we highlight the remaining challenges from two perspectives. First, the purification based method requires retraining the classifier on purified samples which introduces additional computation overhead. Moreover, in a more realistic scenario, testing samples arrives in a streaming fashion and adversarial samples are not isolated from clean samples. These challenges motivates us to explore dynamically update model upon observing testing samples. We proposed a test-time purified self-training strategy to achieve this objective. Adaptive thresholding and feature distribution alignment are introduced to improve the robustness of self-training. Extensive results on different adversarial attacks suggest the proposed method is complementary to purification based method in handling continually changing adversarial attacks on the testing data stream.
\end{abstract}


\end{frontmatter}


\section{Introduction}
\label{sec:intro}

The deployment of 3D point cloud deep learning models is often compromised by their vulnerability to adversarial attacks during the inference stage. This challenge has spurred numerous efforts to enhance the adversarial robustness of these models, such as through adversarial training~\cite{Dong_2020_CVPR}. Although progress has been made in this area, adversarial training has a significant limitation that it requires generating adversarial examples during the training process. Research suggests that adversarial training has poor transferability and may fail when faced with new, unforeseen adversarial attacks. This shortcoming has driven the development of the purification paradigm, which does not depend on specific adversarial attacks for training~\cite{nie2022DiffPure,wu2021ifdefense,sun2023critical,zhang2023ada3diff}. Purification methods typically utilize a generative model to transform input data into a cleaner version, thereby enhancing robustness. For instance, PointDP~\cite{sun2023critical} uses a conditional diffusion model to simultaneously guide generation based on the attacked input point cloud and remove potential adversarial noise. PointDP has demonstrated its ability to improve adversarial robustness across a wide range of attacks without prior exposure to them during training, achieving impressive performance in defending against various adversarial threats.

\begin{figure*}[!htb]
    \centering
    \includegraphics[width=0.9\linewidth]{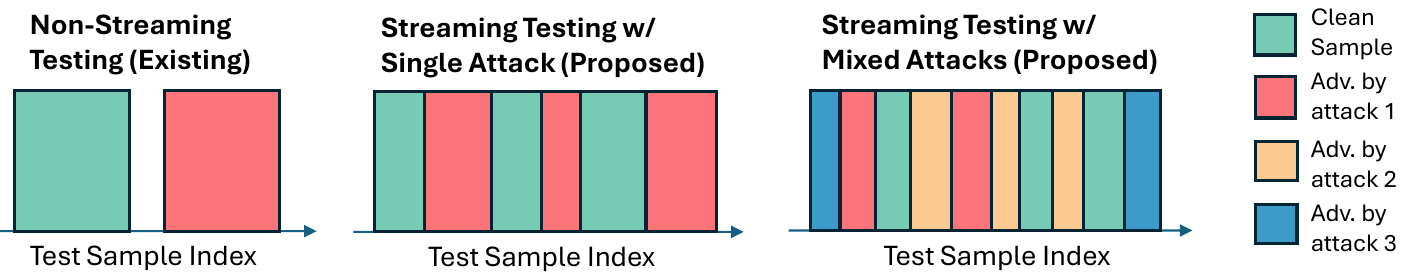}
    \caption{Illustration of the proposed test-time adversarial training protocol. Existing evaluation protocol following a non-stream fashion by separately testing on clean and adversarial samples. Our proposed protocol follows streaming testing with a single type or multiple types~(mixed) of attacks.}
    \label{fig:teaser}
    \vspace{-0.5cm}
\end{figure*}

Despite the success of purifying testing samples to improve robustness, several significant weaknesses remain. First, using a diffusion model to purify testing samples is not specifically tailored to particular types of adversarial attacks. While it is claimed to be robust against unknown adversarial attacks, we believe a more effective defense could be achieved by enhancing the model's robustness with more specific adversarial attacks—information that is only available upon encountering attacked samples at the inference stage. Furthermore, PointDP requires training the prediction model (the point cloud backbone network) on samples generated by the diffusion model. Although this approach improves adversarial robustness, the additional training step may become infeasible when training data is not readily available or when the computational cost of retraining a model is prohibitively expensive.

In light of these challenges, we propose an inference-stage adaptation approach to dynamically adjust the model to emerging adversarial attacks. Rather than anticipating potential adversarial attacks during training, our method focuses on updating the model weights in response to testing data that may contain adversarial examples. To achieve this, we introduce a novel self-training (ST) procedure that builds on existing purification models. This procedure involves making predictions on unlabeled testing samples, and using high-confidence predictions, known as pseudo labels, for further model training.

While traditional self-training approaches for test-time adversarial training have shown effectiveness in defending against attacks~\cite{su2024revisiting}, they often rely on a single fixed threshold~\cite{sohn2020fixmatch}. Given that models may bias toward certain classes, we believe a class-dependent threshold is more suitable for generating high-quality pseudo labels. To this end, we introduce an adaptive threshold for self-training on adversarial testing data streams. Recognizing that self-training alone can be prone to incorrect pseudo labels~\cite{su2022revisiting}, we further incorporate a feature distribution alignment regularization to enhance the robustness of the self-training process. We demonstrate that our proposed self-training procedure is complementary to adversarial purification methods, and their combination leads to significant improvements in adversarial robustness. Therefore, we term the proposed method as Purified Self-Training~(\textbf{PST}).

Finally, we argue that the unique ability of self-training to adapt to unseen adversarial attacks allows us to explore a more realistic and challenging evaluation protocol. This protocol involves testing data arriving in a stream, consisting of both clean samples and adversarial samples generated through diverse attacking methods, as illustrated in Fig.~\ref{fig:teaser}. Such a protocol closely mimics real-world attacks that a hacker might deploy against a discriminative model. To evaluate this, we simulate a testing data stream with continually evolving adversarial attack methods and extensively benchmark our approach against existing purification methods using two variants of the proposed evaluation protocols.

To summarize, we make the following contributions in this work.

\begin{itemize}
   
    \item We first investigate the limitations of state-of-the-art purification methods for improving 3D adversarial robustness and introduce a self-training procedure to adapt models to unseen and ever-changing adversarial attacks at the inference stage.
    \item We further introduce a distribution alignment and adaptive thresholding mechanism to enhance the robustness of self-training procedure.
    \item Finally, we present a novel inference stage adversarial training evaluation protocol which simulates the realistic adversarially attacked testing data stream. Our proposed method demonstrates superior results on the realistic adversarial attacked testing data stream.
\end{itemize}

\section{Related Work}

\subsection{3D Point Cloud Deep Learning}

Early efforts in 3D deep learning primarily focused on adapting 2D models to classify point clouds by utilizing voxelization and 3D convolutional networks. Notable approaches include VoxNet~\cite{Daniel2015voxnet}, SegCloud~\cite{Tchapmi2017segcloud}, and MultiCNN~\cite{Su2015MultiCNN}. However, these methods encountered challenges related to high memory consumption and slow computation, largely due to the inherent sparsity of typical 3D scans. To address these issues, OctNet~\cite{Riegler2017Octnet} introduced the Octree structure, which efficiently handles sparse 3D data, reducing memory usage and improving computation speed. Additionally, generalized sparse convolutions proposed by Choy et al.~\cite{Choy20194DSC} further mitigated memory and computation challenges associated with sparse 3D data.

PointNet~\cite{qi2017pointnet} marked a significant advancement as the first work to directly process raw point clouds without voxelization, leveraging symmetric functions to aggregate features from individual points. Its extension, PointNet\texttt{++}\cite{qi2017pointnetplus}, addressed the limitations of PointNet by capturing local context and hierarchical information. These models have spurred a variety of applications and further developments in 3D deep learning, leading to the creation of models such as PointCNN\cite{li2018pointcnn}, which is specifically designed for point cloud data. Other notable contributions include KPCov~\cite{thomas2019kpconv}, which introduced a novel point convolution technique operating directly on point clouds without intermediate representations, and Dynamic Graph CNN (DGCNN)~\cite{wang2019dynamic}, which leverages local geometric structure through k-nearest neighbor graph construction for each point.

In more recent developments, transformer architectures have garnered attention for 3D point cloud analysis tasks~\cite{lu2022transformers, lahoud20223d_transformers}. Point Transformer~\cite{zhao2021point} applies self-attention within each data point's local neighborhood, while 3DCTN~\cite{lu20223dctn} combines graph convolution layers with transformers, effectively balancing local and global feature learning. LFT-Net~\cite{gao2022lft} introduces self-attention mechanisms specifically for point cloud features, along with a Trans-pooling layer to reduce feature dimensionality. Various approaches, such as the work by Engel et al.\cite{engel2021point}, utilize local-global attention to capture geometric relations, while CpT\cite{kaul2021cpt} employs a dynamic point cloud graph within transformer layers. Furthermore, Point-Voxel Transformer (PVT)~\cite{zhang2021pvt} integrates both voxel-based and point-based transformers for enhanced feature extraction.

\subsection{Adversarial Attacks \& Defense on 3D Point Cloud}

Adversarial attacks, designed to mislead Deep Neural Networks (DNNs) into making incorrect decisions, have been extensively studied in various 2D tasks~\cite{carlini2017evaluating, madry2019deepadv, kurakin2017adversarial, xie2017adversarial, yang2020patchattack}. Notable methods, such as the c\&w attack~\cite{carlini2017evaluating} and Projected Gradient Descent (PGD)\cite{madry2019deepadv}, have been widely explored. However, their direct application to 3D tasks is challenging due to structural differences between 2D and 3D data. Pioneering work by Xiang et al.\cite{xiang2019generating} extended the c\&w attack to 3D point cloud classification, while Wen et al.\cite{wen2020geometryaware} refined the c\&w attack's loss function to achieve smaller perturbations. Tsai et al.\cite{Tsai2020AAAI} incorporated KNN distance into the loss function, enabling the generation of adversarial point clouds that can deceive 3D classifiers in the physical world. Additionally, black-box methods, such as those proposed by Hamdi et al.\cite{Hamdi2020AdvPC} and Zhou et al.\cite{Zhou2020lggan}, have shown enhanced transferability and practicality by using GANs to generate adversarial examples.

When it comes to defending against adversarial attacks on 3D point clouds, several strategies have been developed. Zhou et al.\cite{zhou2019dupnet} introduced defenses like Statistical Outlier Removal (SOR) and Simple Random Sampling (SRS) to eliminate certain points from the point cloud. Dong et al.\cite{Dong_2020_CVPR} proposed defense strategies that involve input transformation and adversarial detection. Liu et al.\cite{liu2019extending} explored the extension of 2D countermeasures to defend against basic attacks like FGSM \cite{goodfellow2015explaining} on point cloud data. Recent advancements have focused on sophisticated purification methods~\cite{wu2021ifdefense, Li_2022_CVPR, sun2023critical, zhang2023ada3diff}. For instance, PointDP~\cite{sun2023critical} utilized diffusion models as a defense mechanism against 3D adversarial attacks. Robustness certification against adversarial attacks has also gained significant attention, with notable contributions from studies like~\cite{Lorenz2021robustcert, S_2022_CVPR, liu2021pointguard}. For example, Liu et al.~\cite{liu2021pointguard} conducted robustness certification for point cloud recognition, considering a threat model that accounts for the number of modified points.

\subsection{Test-Time Training}

In scenarios where models deployed in target domains must autonomously adapt to new environments without access to source domain data, the need for adaptation to arbitrary, unknown target domains with low inference latency has led to the emergence of test-time training/adaptation (TTT/TTA)~\cite{YuSun2019TestTimeTW, wang2020tent}. TTT is typically implemented through the three primary paradigms, self-supervised learning, self-training and distribution alignment. Self-supervised learning on testing data facilitates domain adaptation without relying on semantic information~\cite{YuSun2019TestTimeTW, YuejiangLiu2021TTTWD}. Self-training reinforces the model’s predictions on unlabeled data and has proven effective for TTT~\cite{wang2020tent, chen2022contrastive, liang2020we, goyaltest2022}. Distribution alignment adjusts model weights to align features with the source domain’s distribution~\cite{su2022revisiting, YuejiangLiu2021TTTWD}.
Despite ongoing efforts to refine TTT methods, the certification of TTT's robustness remains an area that has not been fully explored. Recent studies have begun to investigate the robustness of TTT under conditions of target domain distribution shifts over time~\cite{wang2022continual, gongnote2022}. Further exploration has been conducted in scenarios involving small batch sizes and imbalanced classes~\cite{niu2023towards}. Moreover, it has been revealed that TTT is particularly vulnerable in open-world scenarios where the testing data includes strong out-of-distribution (OOD) samples. To address this, Li et al.\cite{li2023robustness} introduced enhancements to the robustness of open-world test-time training through self-training with a dynamically expanded prototype pool. Another approach by Su et al.\cite{su2023realworld} proposes a principled method to simulate more diverse real-world challenges, developing a self-training-based method with balanced batch normalization to achieve state-of-the-art performance.

In this work, we extend test-time training to improve adversarial robustness at the inference stage. We employ self-training with feature distribution alignment regularization, and crucially, we find that this self-training procedure is complementary to existing adversarial purification methods. Combining these approaches results in state-of-the-art adversarial robustness under the new evaluation protocol.

\section{Methodology}
\begin{figure*}
    \centering
    \includegraphics[width=0.99\linewidth]{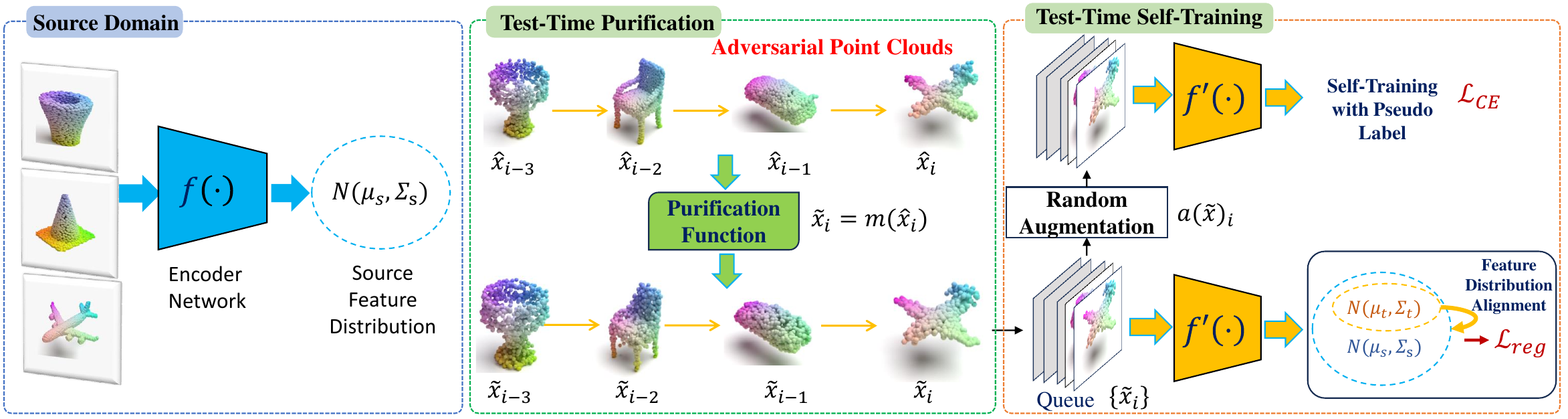}
    \caption{Illustration of test-time purified self-training to improve adversarial robustness. On the source domain, we first collect feature distribution for clean samples. At test-time, we first purify testing samples through off-the-shelf purification method. Self-training is applied to adapt the classification model to unseen adversarial attacks.}
    \label{fig:pipeline}
    \vspace{-0.5cm}
\end{figure*}

To formally define the task, we first denote the testing data stream as 
$ \mathcal{D}_t = \{ (x_i, \bar{y}_i) \mid i = 1, \dots, N_{te} \} $
 where $\bar{y}_i\in\{1\cdots, K\}$ refers to the unknown ground-truth label. The source domain data is available for offline pre-processing, denoted as $ \mathcal{D}_s = \{ (x_i, y_i) \mid i = 1, \dots, N_{tr} \} $
. We further denote a classification model $h(x)\in[0,1]^K$ pre-trained on clean training data and the penultimate layer representation $f(x)\in\mathbbm{R}^{D}$, parameterized by $\Theta$. The attacker may optionally attack the testing sample $x_i$ by maximizing the loss function. The attacking methods could shift over time on the testing data stream. We aim to develop a self-training approach to improve the robustness against shifting adversarial attacks. An overview of the pipeline is presented in Fig.~\ref{fig:pipeline}

\subsection{Review of Self-Training}

Traditional approach towards improving adversarial robustness requires extensive adversarial training~\cite{li2022improving} and specific adversarial attack methods are considered during the training stage. However, the adversarial trained model may fail to generalize to unseen adversarial attacks after deployment. We propose a test-time adaptation approach to improve robustness to unseen attacks. Crucially, we first apply a random augmentation $\mathit{a}(x_i)$ to the input sample and infer the posterior $h(\mathit{a}(x_i))$. We could further obtain the pseudo label $\hat{y}_i$ if the highest probability is above a threshold.
The pseudo labels above the confidence level $\tau$ are further used for updating the network with self-training cross-entropy loss.

\begin{equation}
\begin{split}
    \hat{y}_i&=\arg\max_k h_k(x_i),\\
    \mathcal{L}_{CE}&=\sum_i\sum_k - \mathbbm{1}(\hat{y}_i=k)\mathbbm{1}(\max_k h_k(x_i)>\tau)\log h_k(a(x_i))
\end{split}
\end{equation}

Self-training has been demonstrated to improve robustness against adversarial attacks upon observing attacked testing samples~\cite{su2024revisiting}. Nevertheless, a fixed threshold is required to filter out high confidence predictions for self-training. Identifying the optimal threshold is not a trivial task as no labeled data is available as the validation set. Moreover, the optimal threshold may vary across categories since the model may bias towards certain classes. Motivated by the challenge of specifying a fixed threshold, we propose an adaptive thresholding mechanism for self-training.
\vspace{-0.2cm}

\subsection{Adaptive Thresholding for Self-Training}

There are two concerns for the adaptive threshold. First, the threshold should consider the average model confidence. Moreover, the threshold should be taken into consideration the class distribution. In the following section, we introduce the derivation of adaptive threshold from these two perspectives.

\noindent\textbf{Dynamic Global Confidence}: We characterize the global model confidence by measuring the average confidence. As test-time training goes on, we expect the global confidence to increase. When the attacks changes, we could expect the global confidence to experience a sudden drop. Therefore, such a dynamic global confidence should be estimated from a limited history of testing samples. We denote a queue storing the probabilistic predictions of seen testing samples $\mathcal{Q}=\{x_i\}_{i=1\cdots N_Q}$. The global confidence is defined as follows.

\begin{equation}
    \tau_g = \frac{1}{N_Q}\sum_{x_i\in\mathcal{Q}} \max_kh_k(x_i)
\end{equation}

\noindent\textbf{Class-Aware Thresholding}: The threshold for predicting pseudo labels should be aware of each individual class due to the potential bias of classifier. We follow the practice of distribution alignment~\cite{berthelot2019remixmatch} that encourages the distribution predicted on unlabeled data $\tilde{p}$ to match a prior distribution $p$, which is assumed to be uniform throughout this work. As the predicted distribution is affected by the specific adversarial attacks that may shift over time, we use the average prediction within the testing sample queue $\mathcal{Q}$ to approximate the distribution. We further use the approximated distribution to renormalize the posterior of each testing sample as follows where $Normalize$ refers to a L1 normalization operation.

\begin{equation}
    \tilde{p}=\frac{1}{N_Q}\sum_{x_i\in\mathcal{Q}}h(x_i),\quad \tilde{q}_i=Normalize (h(x_i)\cdot\frac{p}{\tilde{p}})
\end{equation}

The final self-training loss employs the normalized posterior $\tilde{q}$ and the adaptive global confidence for model update.

\begin{equation}
\resizebox{0.99\linewidth}{!}
{$
    \mathcal{L}_{CE}=
    \sum\limits_{ik} - \mathbbm{1}(\arg\max_k \tilde{q}_k(x_i)=k)\mathbbm{1}(\max_k \tilde{q}_k(x_i)>\tau_g)\log h_k(a(x_i))
    $}
\end{equation}

\noindent\textbf{Regularization for Self-Training}:
The above self-training procedure is sensitive to erroneous pseudo labels, a.k.a. confirmation bias~\cite{arazo2020pseudo}. This issue is often less severe in semi-supervised learning due to the existence of labeled loss, while being more challenging during test-time training where labeled data is absent. To mitigate the impact of incorrect pseudo labels, we further introduce a feature distribution alignment regularization~\cite{su2022revisiting} to stabilize self-training. Specifically, we first model the feature distribution in the source domain as a multi-variate Gaussian distribution following Eq.~\ref{eq:source_dist}.

\begin{equation}\label{eq:source_dist}
    \mu_s=\frac{1}{N_{tr}}\sum_{x_i\in\mathcal{D}_s} f(x_i),\quad \Sigma_s=\frac{1}{N_{tr}}\sum_{x_i\in\mathcal{D}_s}(f(x_i)-\mu_s)^\top(f(x_i)-\mu_s)
\end{equation}

The same distribution on the testing data can be inferred and forced to match the source domain distribution. To efficiently infer the distribution on the testing data, we adopt the iterative update strategy~\cite{su2022revisiting} as follows, where $t$ refers to the batch index and $\mathcal{B}^t$ refers to the testing samples at $t$-th batch.

\begin{equation}
\begin{split}
    &\mu^t_t=\mu^{t-1}_t+\sigma^t,\; 
    \sigma^t=\frac{1}{N^t}\sum_{x_i\in\mathcal{B}^t}(f(x_i)-\mu^{t-1}),\quad N^t=N^{t-1}+|\mathcal{B}^t|,\\
     &\Sigma^t_t = \Sigma^{t-1}_t+\sum_{x_i\in\mathcal{B}^t}((f(x_i)-\mu^{t-1})^\top(f(x_i)-\mu^{t-1})-\Sigma^{t-1})-\sigma^{t\top}\sigma^t
\end{split}
\end{equation}

The feature distribution alignment regularization is realised as minimizing the KL-Divergence between the distributions estimated from the training and testing sets. A closed-form solution to the KL-Divergence exists and can be efficiently optimized by gradient descent.

\begin{equation}
    \mathcal{L}_{reg}=KL(\mathcal{N}(\mu_s,\Sigma_s)||\mathcal{N}(\mu_t,\Sigma_t))
\end{equation}

\subsection{Purification for Adversarial Robustness}

Purifying adversarial testing samples is proven a viable way to improve adversarial robustness~\cite{wu2021ifdefense,sun2023critical,zhang2023ada3diff}. Purification is often achieved by fitting an implicit surface~\cite{wu2021ifdefense} or conditional generative model~\cite{sun2023critical,zhang2023ada3diff}. To achieve competitive performance, classifier must be trained on purified training samples to prevent the negative impact of distribution shift. Such a practice restricts the application of purification based method as re-training a large backbone model with purified training samples is computationally expensive. In this work, we combine test-time purification with test-time training and demonstrate that the distribution gap can be well mitigated by test-time training, leading to substantially improved adversarial robustness. Specially, we denote a purification function as $\tilde{x}_i = m(x_i)$. We shall use the purified sample $\tilde{x}_i$ as the input to the self-training procedure introduced above. We demonstrate that the proposed self-training approach is agnostic to the exact implementation of purification and the details of the two purification methods evaluated are presented in the supplementary. Notable, we do not require training the classification model $h(x)$ on the purified training samples to achieve improved robustness.

\vspace{-0.2cm}

\subsection{Purified Self-Training Algorithm}

We present the overall algorithm for the test-time purified self-training in Alg.~\ref{alg:main}. For each minibatch on the test data stream, adversarial samples $\hat{x}$ are first generated via off-the-shelf attack methods. To recognize adversarial testing sample, we first apply purification, followed by data augmentation. The pseudo label is predicted via class aware thresholding. After each minibatch, we update model weights via gradient descent followed by updating the global confidence.

\begin{algorithm}
\caption{Test-Time Purified Self-Training Algorithm }\label{alg:main}

\SetKwInOut{Input}{Input}
\SetKwInOut{Return}{Return}
\Input{Testing sample batch $\mathcal{B}^t = \{x_i\}_{i=1}^{N_B}$, loss weight $\lambda$, pre-trained model weights $\Theta_0$, adversarial attack budget $\epsilon$}
\Return{Model weights $\Theta$}

\textcolor{gray}{\# Inference Stage:}\\
\For{$x_i \leftarrow 1$ \KwTo $N_B$}
{
    Generate adversarial attacks $\hat{x}_i = \arg\max_{||\hat{x}_i - x_i|| \leq \epsilon} \mathcal{L}_{CE}$\;
    Purify testing sample $\tilde{x}_i = m(\hat{x}_i)$\;
    Apply augmentation $a(\tilde{x}_i)$\;
    Predict pseudo label $\hat{y} = \arg\max_k h(\tilde{x}_i)$\;
    Calculate cross-entropy loss $\mathcal{L}_{CE}$ and feature distribution loss $\mathcal{L}_{reg}$\;
}
\textcolor{gray}{\# Gradient Descent Update:}\\
Update $\Theta = \Theta - \alpha (\nabla \mathcal{L}_{CE} + \lambda \nabla \mathcal{L}_{reg})$\;

\textcolor{gray}{\# Update Global Confidence:}\\
Update global confidence by $\tau_g = \frac{1}{N_Q} \sum_{x_i \in \mathcal{Q}} \max_k h_k(\tilde{x}_i)$\;

\end{algorithm}
%

\vspace{-0.5cm}

\subsection{Simulating Testing Data Stream}

In this section, we present two algorithms to introduce the method to simulate testing data stream with single attack and mixed attacks.

\noindent\textbf{Single Attack}: We simulate testing data stream by first generating adversarial attacks $\{x^{adv}\}$ against the frozen source model. Adversarial attacked samples are further mixed with clean samples $\{x^{clean}\}$, resulting in $\{x^{mix}\}=\{x^{adv}\}\bigcup\{x^{clean}\}$. Finally, we randomly shuffle the the mixed samples as the testing data stream.

\noindent\textbf{Mixed Attacks}: We simulate testing stream by generating four types of adversarial attacks, $\{x^{pgd}\}$, $\{x^{cw}\}$, $\{x^{advpc}\}$ and $\{x^{si-adv}\}$. We mixed up the clean $\{x^{clean}\}$ and adversarial attacks to form the testing stream with details given in the following algorithm.

\begin{algorithm}
\caption{Generate Mixed Attack Data}
\SetKwInOut{Input}{Input}
\SetKwInOut{Output}{Output}

\Input{Batch size $B$, Clean data $\{x^{clean}\}$, PGD attack data $\{x^{pgd}\}$, CW attack data $\{x^{cw}\}$, AdvPC attack data $\{x^{advpc}\}$, SI-Adv attack data $\{x^{si-adv}\}$, Data size $N$}

\Output{Mix attack data ${x^{mix}}$}
\SetAlgoLined

\BlankLine
Initialize an empty list $MixData$ to store mixed data;
Initialize a counter $batch\_id = 0$ to alternate between different types of attack data;

\For{$i$ from 0 to $N$ with step size $B$}
{
    Extract a batch of data with index range $[i, i + B)$\ based on $batch\_id$ from:
    \BlankLine
    \Switch{$batch\_id \mod 5$}{
        \Case{0}{clean data $\{x^{clean}\}$}
        \Case{1}{PGD attacked data $\{x^{pgd}\}$}
        \Case{2}{CW attacked data $\{x^{cw}\}$}
        \Case{3}{AdvPC attacked data $\{x^{advpc}\}$}
        \Case{4}{SI-Adv attacked data $\{x^{si-adv}\}$}
    }
}

\Return{Mixed attack data $\{x^{mix}\}$};
\end{algorithm}

\section{Experiment}

\subsection{Experiment Settings}
\noindent\textbf{Dataset}: We evaluate on two widely adopted datasets, ModelNet40~\cite{wu20153d} and ScanObjectNN~\cite{uy-scanobjectnn-iccv19} for the test-time adversarial robustness. ModelNet40 comprises 12,311 CAD models spanning 40 different categories, with 9,843 samples allocated for model training and 2,468 samples for model testing. Following established practices~\cite{qi2017pointnet}, we uniformly sample 1,024 points from the surface of each sample in the dataset and scale them to fit within a unit cube. We also utilize the ScanObjectNN Dataset~\cite{uy-scanobjectnn-iccv19}, composed of 2,902 objects spanning 15 categories derived from real scanning data. Following established practices, we divided the dataset into 9,843 training samples and 2,468 testing samples, with each object consisting of 2048 points. Additionally, to validate the superiority of our method, we employed the most challenging perturbation variant data with background information, i.e., \textit{PB\_T50\_RS}.

To obtain a pre-trained classification model, we apply data augmentation including translation and shuffling on clean training samples. At testing stage, we apply random perturbations, scaling, cropping, rotation, translation, and global affine transformations to testing sample (both clean and adversarial samples). These augmentations contribute to enhancing the model's robustness and generalization capabilities.

\noindent\textbf{Backbone Network}: We conducted experiments on three distinct architectures of 3D recognition models, namely PointNet\texttt{++}~\cite{qi2017pointnetplus}, PCT~\cite{guo2021pct}, CurveNet~\cite{xiang2021walk}, to assess the generality of our proposed method. Evaluations on additional backbone networks are deferred to the supplementary material.

\noindent\textbf{Hyperparameters}: We adopt the following hyperparameters. The weight applied to regularization term is $\lambda=0.5$. We apply batchsize $32$ and learning rate $1e-3$ for self-training. The queue size is configured to 512. We adopt Stochastic Gradient Descent (SGD) as our optimizer. 

{We introduce several data augmentation techniques for point clouds to enhance the robustness and diversity of training data. For random perturbation, Gaussian noise with a mean of 0 and a standard deviation of 0.01 is added to the three dimensions of the point cloud. Scaling involves applying random factors ranging from 0.6 to 1.4 to each dimension of the point cloud individually. Shearing is implemented by applying a shear matrix that proportionally moves points on one axis to another, with a shear angle of 0.5 degrees. Rotation entails rotating the point cloud around each axis by 1 degree. Translation involves scaling each point of the point cloud individually by random factors ranging from 0.6 to 1.4 and adding random offsets in each dimension ranging from -0.2 to 0.2. For global affine transformation, a random affine transformation matrix is applied, consisting of the identity matrix with random noise sampled from a Gaussian distribution with a mean of 0 and a standard deviation of 0.1. For each point cloud sample, we randomly select a combination of the three data augmentation methods mentioned above.}
      
\noindent\textbf{Adversarial Attacks}:

In our study, we directly attack point cloud classification models using widely adopted adversarial techniques, including PGD~\cite{madry2019deepadv}, C\&W~\cite{carlini2017evaluating}, imperceptible SI\_Adv~\cite{huang2022siadv}, and specifically designed attacks for point cloud models such as KNN~\cite{Tsai2020AAAI}, and AdvPC~\cite{Hamdi2020AdvPC}. In addition, we also employ strong adaptive attacks. Following the methodology of PointDP~\cite{sun2023critical}, as point cloud diffusion models rely on latent features, we perform PGD and AdvPC attacks targeted at latent features when PointDP purification is included. Additionally, we evaluate gradient estimation attacks on the reverse generation process of diffusion models, specifically the method proposed by DiffPure\cite{nie2022DiffPure}. We also assess AutoAttack~\cite{croce2020reliable} based on PGD attacks. 
For black-box attacks, we evaluate attacks on the diffusion generation results, including SPSA~\cite{uesato2018adversarial} and Nattack~\cite{li2019nattack}, following the recommendations of Carlini et al.~\cite{carlini2019evaluating}. We use \( l_\infty \) and \( l_2 \) norms as metrics for measuring the magnitude of perturbations.
For SI-Adv, we adhere to the suggested approach, constraining the perturbations within an \( l_\infty \) norm ball with a radius of 0.16. For other attacks, we follow the practices of PointDP and~\cite{sun2021adversarially}, utilizing \( l_\infty \) norm balls with a radius of 0.05 and  \( l_2 \)  norm balls with a radius of 1.25 to constrain perturbations. The number of attack iterations is set to 200 to ensure sufficient adversarial effectiveness. 

\noindent\textbf{Defense Methods}: We evaluate two off-the-shelf purification methods. \textbf{IF-Defense}~\cite{wu2021ifdefense} proposed a framework based on implicit functions to learn the
restoration of clean point clouds from adversarially perturbed point clouds. \textbf{PointDP}~\cite{sun2023critical} employs a pre-trained diffusion model to generate purified point cloud conditioned on input adversarially perturbed one. Finally, we combine each off-the-shelf purification method with the proposed purified self-training~(\textbf{PST}) to further improve the adversarial robustness at inference stage.

\noindent\textbf{Evaluation Protocol}:
We proposed a novel adversarial attack evaluation protocol to benchmark existing methods. Specifically, there are two variants of streaming testing data. The first assumes only a single type of adversarial attack is employed to generate adversarial samples while the second assumes multiple types of adversarial attacks are employed sequentially to generate adversarial attacks. For both protocols, the clean and adversarial testing samples are mixed up in the testing data stream. An illustration of the protocols compared with existing non-streaming testing data is presented in Fig.~\ref{fig:teaser}. To simulate the streaming testing data, we first randomly divide the test set into 2 and 5 subsets for \textbf{Single Attack} and \textbf{Mixed Attack} respectively. We attack the subset 2 and 2 to 5 respectively with one of the four adversarial attacks (C\&W, PGD, AdvPC and SI\_Adv) and reserve subset 1 as clean testing samples. All subsets are then randomly shuffled to form the final testing data stream. More details of simulating the testing data stream is deferred to the supplementary.

To properly evaluate the performance on the proposed streaming testing data, we adopt three evaluating metrics. In specific, we calculate the clean accuracy, adversarial accuracy and mixed accuracy. The former two follow the same definition made in previous works~\cite{szegedy2013intriguing} while the mixed accuracy is the average accuracy over both clean and adversarial testing samples. 

\begin{table}[!tbp]
  \centering
  \caption{Evaluation of robustness against white-box attacks on ModelNet40 dataset under Single Attack protocol with five types of adversarial attacks.}
  \resizebox{0.99\linewidth}{!}{
\setlength{\tabcolsep}{3pt} 
\begin{tabular}{cllcccc|cccc}
\hline
 &  &  & \multicolumn{4}{c|}{\( l_\infty \)} & \multicolumn{4}{c}{\( l_2 \)} \\ \cline{4-11} 
 &  &  & C\&W & PGD & AdvPC & SI\_Adv & C\&W & PGD & AdvPC & KNN \\ \hline
\multirow{12}{*}{\begin{sideways}CurveNet\end{sideways}} & \multirow{3}{*}{PointDP~\cite{sun2023critical}} & Clean Acc & 80.4 & 80.4 & 80.4 & 80.4 & 80.4 & 80.4 & 80.4 & 80.4 \\
 &  & Adv Acc & 77.6 & 69.9 & 63.2 & 77.2 & 75.1 & 57.9 & 62.8 & 79.2 \\
 &  & Mixed Acc & 79.0 & 75.2 & 71.8 & 78.8 & 77.8 & 69.2 & 71.6 & 79.8 \\ \cline{2-11} 
 & \multirow{3}{*}{PointDP+PST(Ours)} & Clean Acc & \textbf{84.5} & \textbf{84.1} & \textbf{83.6} & \textbf{84.0} & \textbf{84.8} & \textbf{83.8} & \textbf{83.7} & \textbf{85.1} \\
 &  & Adv Acc & \textbf{82.3} & \textbf{77.4} & \textbf{75.3} & \textbf{82.6} & \textbf{82.7} & \textbf{72.7} & \textbf{74.8} & \textbf{84.3} \\
 &  & Mixed Acc & \textbf{83.4} & \textbf{80.7} & \textbf{79.5} & \textbf{83.3} & \textbf{83.8} & \textbf{78.3} & \textbf{79.3} & \textbf{84.7} \\ \cline{2-11} 
 & \multirow{3}{*}{IF-Def~\cite{wu2021ifdefense}} & Clean Acc & 78.4 & 78.9 & 80.1 & 78.4 & 79.1 & 78.4 & 78.2 & 78.4 \\
 &  & Adv Acc & 83.1 & 54.6 & 54.2 & 82.2 & 82.3 & 69.6 & 53.9 & 84.8 \\
 &  & Mixed Acc & 80.7 & 66.8 & 67.1 & 80.3 & 80.7 & 74.0 & 66.1 & 81.7 \\ \cline{2-11} 
 & \multirow{3}{*}{IF-Def+PST(Ours)} & Clean Acc & \textbf{82.5} & \textbf{80.5} & \textbf{80.7} & \textbf{81.2} & \textbf{80.6} & \textbf{79.8} & \textbf{79.8} & \textbf{81.7} \\
 &  & Adv Acc & \textbf{84.3} & \textbf{67.7} & \textbf{67.6} & \textbf{85.6} & \textbf{84.1} & \textbf{76.5} & \textbf{67.0} & \textbf{86.7} \\
 &  & Mixed Acc & \textbf{82.5} & \textbf{74.1} & \textbf{74.2} & \textbf{83.4} & \textbf{82.3} & \textbf{78.2} & \textbf{73.4} & \textbf{84.2} \\ \hline\hline
\multirow{12}{*}{\begin{sideways}PointNet++\end{sideways}} & \multirow{3}{*}{PointDP~\cite{sun2023critical}} & Clean Acc & 78.1 & 78.1 & 78.1 & 78.1 & 78.1 & 78.1 & 78.1 & 78.1 \\
 &  & Adv Acc & 74.4 & 66.9 & 62.2 & 71.6 & 67.4 & 69.7 & 62.0 & 76.9 \\
 &  & Mixed Acc & 76.2 & 72.5 & 70.2 & 74.8 & 72.7 & 73.9 & 70.1 & 77.5 \\ \cline{2-11} 
 & \multirow{3}{*}{PointDP+PST(Ours)} & Clean Acc & \textbf{83.6} & \textbf{83.3} & \textbf{83.4} & \textbf{83.5} & \textbf{81.9} & \textbf{83.6} & \textbf{83.6} & \textbf{83.8} \\
 &  & Adv Acc & \textbf{81.3} & \textbf{79.6} & \textbf{76.0} & \textbf{81.7} & \textbf{80.0} & \textbf{78.8} & \textbf{76.0} & \textbf{82.9} \\
 &  & Mixed Acc & \textbf{82.5} & \textbf{81.4} & \textbf{79.7} & \textbf{82.6} & \textbf{81.0} & \textbf{81.2} & \textbf{79.8} & \textbf{83.4} \\ \cline{2-11} 
 & \multirow{3}{*}{IF-Def~\cite{wu2021ifdefense}} & Clean Acc & 79.8 & 79.2 & 79.2 & 79.0 & 79.2 & 79.1 & 79.2 & 79.3 \\
 &  & Adv Acc & 83.4 & 73.0 & 70.3 & 83.5 & 81.2 & 77.2 & 71.1 & 83.9 \\
 &  & Mixed Acc & 81.6 & 76.1 & 74.8 & 81.2 & 80.2 & 78.2 & 75.2 & 81.6 \\ \cline{2-11} 
 & \multirow{3}{*}{IF-Def+PST(Ours)} & Clean Acc & \textbf{82.4} & \textbf{81.2} & \textbf{80.4} & \textbf{81.4} & \textbf{80.7} & \textbf{81.7} & \textbf{81.3} & \textbf{81.0} \\
 &  & Adv Acc & \textbf{84.8} & \textbf{77.3} & \textbf{76.5} & \textbf{85.9} & \textbf{83.0} & \textbf{80.4} & \textbf{77.3} & \textbf{84.7} \\
 &  & Mixed Acc & \textbf{83.6} & \textbf{79.3} & \textbf{78.5} & \textbf{83.6} & \textbf{81.9} & \textbf{81.0} & \textbf{79.3} & \textbf{82.8} \\ \hline\hline
\multirow{12}{*}{{\begin{sideways}PCT\end{sideways}}} & \multirow{3}{*}{PointDP~\cite{sun2023critical}} & Clean Acc & 81.2 & 81.2 & 81.2 & 81.2 & 81.2 & 81.2 & 81.2 & 81.2 \\
 &  & Adv Acc & 77.5 & 70.4 & 64.5 & 76.2 & 75.0 & 60.0 & 65.3 & 80.4 \\
 &  & Mixed Acc & 79.3 & 75.8 & 72.8 & 78.7 & 78.1 & 70.6 & 73.2 & 80.8 \\ \cline{2-11} 
 & \multirow{3}{*}{PointDP+PST(Ours)} & Clean Acc & \textbf{82.4} & \textbf{83.8} & \textbf{83.4} & {81.0} & \textbf{83.8} & \textbf{83.3} & \textbf{83.9} & \textbf{84.6} \\
 &  & Adv Acc & \textbf{80.1} & \textbf{75.7} & \textbf{71.5} & \textbf{82.8} & \textbf{80.3} & \textbf{71.0} & \textbf{72.0} & \textbf{83.0} \\
 &  & Mixed Acc & \textbf{81.2} & \textbf{79.7} & \textbf{77.5} & \textbf{82.0} & \textbf{82.1} & \textbf{77.2} & \textbf{78.0} & \textbf{83.8} \\ \cline{2-11} 
 & \multirow{3}{*}{IF-Def~\cite{wu2021ifdefense}} & Clean Acc & \textbf{80.3} & 80.9 & 81.5 & 80.9 & 80.9 & 80.5 & 81.3 & 80.4 \\
     &  & Adv Acc & 83.2 & 60.8 & 60.4 & 84.3 & 83.3 & 70.2 & 59.7 & \textbf{86.6} \\
 &  & Mixed Acc & 81.7 & 70.9 & 71.0 & 82.6 & 82.1 & 75.4 & 70.5 & \textbf{83.5} \\ \cline{2-11} 
 & \multirow{3}{*}{IF-Def+PST(Ours)} & Clean Acc & 80.2 & \textbf{82.0} & \textbf{82.5} & \textbf{82.3} & \textbf{82.5} & \textbf{81.6} & \textbf{81.9} & \textbf{80.4} \\
 &  & Adv Acc & \textbf{83.5} & \textbf{69.5} & \textbf{67.4} & \textbf{85.0} & \textbf{84.5} & \textbf{75.1} & \textbf{69.6} & {84.7} \\
 &  & Mixed Acc & \textbf{81.9} & \textbf{75.7} & \textbf{75.0} & \textbf{83.6} & \textbf{83.6} & \textbf{78.4} & \textbf{75.8} & {82.5} \\ \hline
\end{tabular}
  }  \label{tab:ModelNet40_white}%
\end{table}

\begin{table}[!tbp]
  \centering
  \caption{Evaluation of robustness against white-box attacks on ModelNet40 dataset under Mixed Attack protocol with five types of adversarial attacks.}
  \resizebox{0.99\linewidth}{!}{
\setlength{\tabcolsep}{3pt} 
\begin{tabular}{cllcrrr|crrr}
\hline
\multicolumn{1}{l}{} &  &  & \multicolumn{4}{c|}{\( l_\infty \)} & \multicolumn{4}{c}{\( l_2 \)} \\ \cline{4-11} 
\multicolumn{1}{l}{} &  &  & C\&W & \multicolumn{1}{c}{PGD} & \multicolumn{1}{c}{AdvPC} & \multicolumn{1}{c|}{SI\_Adv} & C\&W & \multicolumn{1}{c}{PGD} & \multicolumn{1}{c}{AdvPC} & \multicolumn{1}{c}{KNN} \\ \hline
\multirow{12}{*}{{\begin{sideways}CurveNet\end{sideways}}} & \multirow{3}{*}{PointDP~\cite{sun2023critical}} & Clean Acc & \multicolumn{4}{c|}{72.3} & \multicolumn{4}{c}{72.3} \\
 &  & Adv Acc & \multicolumn{1}{r}{74.8} & 71.2 & 76.0 & 71.6 & \multicolumn{1}{r}{74.0} & 54.3 & 76.0 & \textbf{76.1} \\
 &  & Mixed Acc & \multicolumn{4}{c|}{73.2} & \multicolumn{4}{c}{70.5} \\ \cline{2-11} 
 & \multirow{3}{*}{PointDP+PST(Ours)} & Clean Acc & \multicolumn{4}{c|}{\textbf{78.2}} & \multicolumn{4}{c}{\textbf{80.4}} \\
 &  & Adv Acc & \multicolumn{1}{r}{\textbf{84.4}} & \textbf{82.4} & \textbf{89.5} & \textbf{73.6} & \multicolumn{1}{r}{\textbf{79.3}} & \textbf{77.5} & \textbf{89.5} & 75.9 \\
 &  & Mixed Acc & \multicolumn{4}{c|}{\textbf{81.6}} & \multicolumn{4}{c}{\textbf{80.5}} \\ \cline{2-11} 
 & \multirow{3}{*}{IF-Def~\cite{wu2021ifdefense}} & Clean Acc & \multicolumn{4}{c|}{81.8} & \multicolumn{4}{c}{82.2} \\
 &  & Adv Acc & \multicolumn{1}{r}{80.9} & 55.7 & 65.1 & \textbf{78.9} & \multicolumn{1}{r}{77.4} & 77.8 & 65.7 & \textbf{80.3} \\
 &  & Mixed Acc & \multicolumn{4}{c|}{72.5} & \multicolumn{4}{c}{76.7} \\ \cline{2-11} 
 & \multirow{3}{*}{IF-Def+PST(Ours)} & Clean Acc & \multicolumn{4}{c|}{\textbf{84.9}} & \multicolumn{4}{c}{\textbf{85.3}} \\
 &  & Adv Acc & \multicolumn{1}{r}{\textbf{79.1}} & \textbf{71.2} & \textbf{83.6} & 77.1 & \multicolumn{1}{r}{\textbf{80.1}} & \textbf{79.7} & \textbf{84.4} & 77.3 \\
 &  & Mixed Acc & \multicolumn{4}{c|}{\textbf{79.2}} & \multicolumn{4}{c}{\textbf{81.4}} \\ \hline\hline
\multirow{12}{*}{{\begin{sideways}PointNet++\end{sideways}}} & \multirow{3}{*}{PointDP~\cite{sun2023critical}} & Clean Acc & \multicolumn{4}{c|}{70.3} & \multicolumn{4}{c}{70.3} \\
 &  & Adv Acc & \multicolumn{1}{r}{72.2} & 70.9 & 74.0 & 67.5 & \multicolumn{1}{r}{65.9} & 68.9 & 74.0 & 69.3 \\
 &  & Mixed Acc & \multicolumn{4}{c|}{71.0} & \multicolumn{4}{c}{69.7} \\ \cline{2-11} 
 & \multirow{3}{*}{PointDP+PST(Ours)} & Clean Acc & \multicolumn{4}{c|}{\textbf{75.0}} & \multicolumn{4}{c}{\textbf{75.8}} \\
 &  & Adv Acc & \multicolumn{1}{r}{\textbf{80.7}} & \textbf{85.4} & \textbf{91.5} & \textbf{74.1} & \multicolumn{1}{r}{\textbf{78.7}} & \textbf{81.5} & \textbf{91.3} & \textbf{75.7} \\
 &  & Mixed Acc & \multicolumn{4}{c|}{\textbf{81.3}} & \multicolumn{4}{c}{\textbf{80.6}} \\ \cline{2-11} 
 & \multirow{3}{*}{IF-Def~\cite{wu2021ifdefense}} & Clean Acc & \multicolumn{4}{c|}{84.0} & \multicolumn{4}{c}{84.6} \\
 &  & Adv Acc & \multicolumn{1}{r}{79.3} & 75.0 & 82.3 & \textbf{79.3} & \multicolumn{1}{r}{74.2} & 83.5 & 79.5 & \textbf{78.5} \\
 &  & Mixed Acc & \multicolumn{4}{c|}{80.0} & \multicolumn{4}{c}{80.1} \\ \cline{2-11} 
 & \multirow{3}{*}{IF-Def+PST(Ours)} & Clean Acc & \multicolumn{4}{c|}{\textbf{84.1}} & \multicolumn{4}{c}{\textbf{83.1}} \\
 &  & Adv Acc & \multicolumn{1}{r}{\textbf{79.7}} & \textbf{84.4} & \textbf{89.9} & 78.1 & \multicolumn{1}{r}{\textbf{78.7}} & \textbf{86.6} & \textbf{88.7} & 78.3 \\
 &  & Mixed Acc & \multicolumn{4}{c|}{\textbf{83.2}} & \multicolumn{4}{c}{\textbf{83.1}} \\ \hline\hline
\multirow{12}{*}{{\begin{sideways}PCT\end{sideways}}} & \multirow{3}{*}{PointDP~\cite{sun2023critical}} & Clean Acc & \multicolumn{4}{c|}{73.9} & \multicolumn{4}{c}{73.9} \\
 &  & Adv Acc & \multicolumn{1}{r}{71.4} & 72.2 & 81.1 & 74.0 & \multicolumn{1}{r}{70.5} & 58.0 & 81.1 & \textbf{78.1} \\
 &  & Mixed Acc & \multicolumn{4}{c|}{74.5} & \multicolumn{4}{c}{72.3} \\ \cline{2-11} 
 & \multirow{3}{*}{PointDP+PST(Ours)} & Clean Acc & \multicolumn{4}{c|}{\textbf{77.6}} & \multicolumn{4}{c}{\textbf{77.6}} \\
 &  & Adv Acc & \multicolumn{1}{r}{\textbf{76.7}} & \textbf{78.7} & \textbf{89.1} & \textbf{75.9} & \multicolumn{1}{r}{\textbf{78.7}} & \textbf{78.7} & \textbf{84.8} & 75.9 \\
 &  & Mixed Acc & \multicolumn{4}{c|}{\textbf{79.6}} & \multicolumn{4}{c}{\textbf{79.1}} \\ \cline{2-11} 
 & \multirow{3}{*}{IF-Def~\cite{wu2021ifdefense}} & Clean Acc & \multicolumn{4}{c|}{\textbf{83.6}} & \multicolumn{4}{c}{82.4} \\
 &  & Adv Acc & \multicolumn{1}{r}{78.5} & 61.6 & 74.2 & \textbf{80.7} & \multicolumn{1}{r}{78.9} & 73.0 & 74.2 & \textbf{79.1} \\
 &  & Mixed Acc & \multicolumn{4}{c|}{75.7} & \multicolumn{4}{c}{77.5} \\ \cline{2-11} 
 & \multirow{3}{*}{IF-Def+PST(Ours)} & Clean Acc & \multicolumn{4}{c|}{{82.3}} & \multicolumn{4}{c}{\textbf{83.9}} \\
 &  & Adv Acc & \multicolumn{1}{r}{\textbf{85.4}} & \textbf{71.2} & \textbf{85.4} & 79.7 & \multicolumn{1}{r}{\textbf{81.7}} & \textbf{78.5} & \textbf{84.2} & 77.9 \\
 &  & Mixed Acc & \multicolumn{4}{c|}{\textbf{79.4}} & \multicolumn{4}{c}{\textbf{81.3}} \\ \hline
\end{tabular}
  }  \label{tab:ModelNet40_white_Mixed}%
\end{table}

\begin{table*}[!tbp]
  \centering
  \caption{Evaluation of robustness against white-box attacks on ScanObjectNN dataset under Single and Mixed Attack protocol with four types of adversarial attacks.}
  \resizebox{0.9\linewidth}{!}{
\setlength{\tabcolsep}{4pt}
\begin{tabular}{cllccccccc|ccccccc}
\hline
\multicolumn{1}{l}{\multirow{2}{*}{}} & \multirow{2}{*}{} & \multirow{2}{*}{} & \multicolumn{7}{c|}{Single Attack} & \multicolumn{7}{c}{Mixed Attack} \\ 
\cline{4-17} 
\multicolumn{1}{l}{} &  &  & \multicolumn{3}{c|}{\( l_\infty \)} & \multicolumn{4}{c|}{\( l_2 \)} & \multicolumn{3}{c|}{\( l_\infty \)} & \multicolumn{4}{c}{\( l_2 \)} \\ \cline{4-17} 
\multicolumn{1}{l}{} &  &  & C\&W & PGD & \multicolumn{1}{c|}{AdvPC} & C\&W & PGD & AdvPC & KNN & C\&W & PGD & \multicolumn{1}{c|}{AdvPC} & C\&W & PGD & AdvPC & KNN \\ \hline 
\multirow{6}{*}{\begin{sideways}CurveNet\end{sideways}} & \multirow{3}{*}{PointDP~\cite{sun2023critical}} & Clean Acc & 45.0 & 45.0 & \multicolumn{1}{c|}{45.0} & 45.0 & 45.0 & 45.0 & 45.0 & \multicolumn{3}{c|}{42.3} & \multicolumn{4}{c}{42.8} \\
 &  & Adv Acc & 34.9 & 23.7 & \multicolumn{1}{c|}{21.5} & 33.2 & 24.1 & 46.8 & 45.8 & 34.4 & 22.8 & \multicolumn{1}{c|}{27.3} & 38.0 & 21.3 & 50.1 & 36.8 \\
 &  & Mixed Acc & 40.0 & 34.9 & \multicolumn{1}{c|}{33.3} & 39.1 & 34.6 & 45.9 & 45.4 & \multicolumn{3}{c|}{31.6} & \multicolumn{4}{c}{37.8} \\ \cline{2-17} 
 & \multirow{3}{*}{PointDP+PST(Ours)} & Clean Acc & \textbf{57.7} & \textbf{58.3} & \multicolumn{1}{c|}{\textbf{58.0}} & \textbf{58.7} & \textbf{58.5} & \textbf{59.7} & \textbf{59.4} & \multicolumn{3}{c|}{\textbf{58.7}} & \multicolumn{4}{c}{\textbf{61.6}} \\
 &  & Adv Acc & \textbf{54.4} & \textbf{44.3} & \multicolumn{1}{c|}{\textbf{41.5}} & \textbf{53.1} & \textbf{44.0} & \textbf{59.8} & \textbf{60.0} & \textbf{54.0} & \textbf{45.7} & \multicolumn{1}{c|}{\textbf{41.0}} & \textbf{54.5} & \textbf{47.9} & \textbf{56.1} & \textbf{50.9} \\
 &  & Mixed Acc & \textbf{56.1} & \textbf{51.3} & \multicolumn{1}{c|}{\textbf{49.8}} & \textbf{55.9} & \textbf{51.3} & \textbf{59.8} & \textbf{59.7} & \multicolumn{3}{c|}{\textbf{49.7}} & \multicolumn{4}{c}{\textbf{54.2}} \\ \hline \hline
\multirow{6}{*}{\begin{sideways}PointNet++\end{sideways}} & \multirow{3}{*}{PointDP~\cite{sun2023critical}} & Clean Acc & 47.4 & 47.4 & \multicolumn{1}{c|}{47.4} & 47.4 & 47.4 & 47.4 & 47.4 & \multicolumn{3}{c|}{42.4} & \multicolumn{4}{c}{57.2} \\
 &  & Adv Acc & 37.4 & 21.4 & \multicolumn{1}{c|}{16.8} & 27.7 & 31.9 & 47.0 & 47.0 & 40.6 & 18.4 & \multicolumn{1}{c|}{18.8} & 27.6 & 21.6 & 53.1 & 44.2 \\
 &  & Mixed Acc & 42.0 & 34.4 & \multicolumn{1}{c|}{32.1} & 27.6 & 39.4 & 47.2 & 47.2 & \multicolumn{3}{c|}{29.8} & \multicolumn{4}{c}{40.7} \\ \cline{2-17} 
 & \multirow{3}{*}{PointDP+PST(Ours)} & Clean Acc & \textbf{54.7} & \textbf{57.4} & \multicolumn{1}{c|}{\textbf{54.9}} & \textbf{56.9} & \textbf{56.2} & \textbf{57.5} & \textbf{57.1} & \multicolumn{3}{c|}{\textbf{54.4}} & \multicolumn{4}{c}{\textbf{62.9}} \\
 &  & Adv Acc & \textbf{48.8} & \textbf{47.0} & \multicolumn{1}{c|}{\textbf{41.7}} & \textbf{48.8} & \textbf{49.5} & \textbf{58.1} & \textbf{57.6} & \textbf{51.1} & \textbf{48.0} & \multicolumn{1}{c|}{\textbf{44.5}} & \textbf{45.3} & \textbf{47.2} & \textbf{53.7} & \textbf{64.9} \\
 &  & Mixed Acc & \textbf{51.7} & \textbf{52.2} & \multicolumn{1}{c|}{\textbf{48.3}} & \textbf{52.9} & \textbf{52.9} & \textbf{57.8} & \textbf{57.4} & \multicolumn{3}{c|}{\textbf{49.4}} & \multicolumn{4}{c}{\textbf{54.8}} \\ \hline \hline
\multirow{6}{*}{\begin{sideways}PCT\end{sideways}} & \multirow{3}{*}{PointDP~\cite{sun2023critical}} & Clean Acc & 47.5 & 47.5 & \multicolumn{1}{c|}{47.5} & 47.5 & 47.5 & 47.5 & 47.5 & \multicolumn{3}{c|}{49.0} & \multicolumn{4}{c}{47.9} \\
 &  & Adv Acc & 39.3 & 24.8 & \multicolumn{1}{c|}{22.9} & 35.5 & 23.1 & 49.7 & 47.0 & 52.3 & 15.6 & \multicolumn{1}{c|}{12.2} & 38.3 & 19.9 & 47.7 & 38.8 \\
 &  & Mixed Acc & 43.7 & 36.2 & \multicolumn{1}{c|}{35.2} & 41.5 & 35.3 & 48.6 & 47.3 & \multicolumn{3}{c|}{32.3} & \multicolumn{4}{c}{38.5} \\ \cline{2-17} 
 & \multirow{3}{*}{PointDP+PST(Ours)} & Clean Acc & \textbf{54.3} & \textbf{53.4} & \multicolumn{1}{c|}{\textbf{54.5}} & \textbf{54.7} & \textbf{54.3} & \textbf{56.0} & \textbf{56.3} & \multicolumn{3}{c|}{\textbf{52.1}} & \multicolumn{4}{c}{\textbf{56.9}} \\
 &  & Adv Acc & \textbf{48.0} & \textbf{40.8} & \multicolumn{1}{c|}{\textbf{39.8}} & \textbf{50.6} & \textbf{38.6} & \textbf{58.4} & \textbf{56.9} & \textbf{71.9} & \textbf{30.3} & \multicolumn{1}{c|}{\textbf{23.1}} & \textbf{53.8} & \textbf{39.4} & \textbf{56.9} & \textbf{54.9} \\
 &  & Mixed Acc & \textbf{51.2} & \textbf{47.1} & \multicolumn{1}{c|}{\textbf{47.2}} & \textbf{52.7} & \textbf{46.5} & \textbf{57.2} & \textbf{56.6} & \multicolumn{3}{c|}{\textbf{44.3}} & \multicolumn{4}{c}{\textbf{52.4}} \\ \hline
\end{tabular}
  }  \label{tab:scanobj_single_mix}%
\end{table*}

\subsection{Results on Adversarial Robustness}

\noindent\textbf{White-box Attacks}: 
We first evaluate the robustness against white-box attacks. Since the model weights are subject to constant update by self-training, we assume the attack only has the access to original model weights for generating adversarial attacks rather than the real-time weights. We first report the results with white-box attacks under the \textbf{Single Attack} protocol.
We report the results with white-box attacks, i.e. C\&W, PGD, AdvPC, SI\_Adv and KNN, in Tab.~\ref{tab:ModelNet40_white} and Tab.~\ref{tab:ModelNet40_white_Mixed} for ModelNet40 dataset and Tab.~\ref{tab:scanobj_single_mix} for ScanObjNN dataset, respectively. We benchmark against two purification methods, i.e. PointDP~(Pt.~DP)~\cite{sun2023critical} and IF-Defense~(IF-Def.)~\cite{wu2021ifdefense}. For each purification method, we combine with purified self-training, denoted as PointDP+PST and IF-Def+PST as out methods. We make the following observations from the results. First, under Single Attack, our method consistently outperforms both PointDP and IF-Defense with a noticeable margin. In particular, our method leads a more significant margin on stronger attacks, e.g. PGD and AdvPC. Second, both PointDP and IF-Defense are susceptible to certain attacks. For example, PointDP performs substantially worse on PGD and AdvPC attacks. In comparison, despite our method utilizes PointDP for purification, we consistently improve the robustness to the more challenging attacks. At the same time, our self-training also benefits the accuracy on clean samples. Finally, the observations are consistent across three backbone networks, covering MLP and transformer based designs. The improvement is even more significant on ScanObjectNN dataset which represents the real-world point cloud recognition task.

\begin{table*}[!tbp]
  \centering
  \caption{Evaluation of robustness against adaptive attacks on ModelNet40 and ScanObjectNN datasets under Single Attack protocol.}
   \resizebox{1\linewidth}{!}{

\setlength{\tabcolsep}{4pt}
\begin{tabular}{clccccccccc|cccccccc}
\hline
\multirow{2}{*}{} & \multirow{2}{*}{} & \multirow{2}{*}{} & \multicolumn{8}{c|}{ModelNet40} & \multicolumn{8}{c}{ScanObjectNN} \\ \cline{4-19} 
 &  &  & \multicolumn{4}{c|}{\( l_\infty \)} & \multicolumn{4}{c|}{\( l_2 \)} & \multicolumn{4}{c|}{\( l_\infty \)} & \multicolumn{4}{c}{\( l_2 \)} \\ \cline{4-19} 
 &  &  & \multicolumn{1}{c|}{\begin{tabular}[c]{@{}c@{}}BPDA-\\ PGD\end{tabular}} & \multicolumn{1}{c|}{\begin{tabular}[c]{@{}c@{}}EOT-\\ AutoAtt.\end{tabular}} & \multicolumn{1}{c|}{\begin{tabular}[c]{@{}c@{}}PGD-\\ Latent\end{tabular}} & \multicolumn{1}{c|}{\begin{tabular}[c]{@{}c@{}}AdvPC-\\ Latent\end{tabular}} & \multicolumn{1}{c|}{\begin{tabular}[c]{@{}c@{}}BPDA-\\ PGD\end{tabular}} & \multicolumn{1}{c|}{\begin{tabular}[c]{@{}c@{}}EOT-\\ AutoAtt.\end{tabular}} & \multicolumn{1}{c|}{\begin{tabular}[c]{@{}c@{}}PGD-\\ Latent\end{tabular}} & \begin{tabular}[c]{@{}c@{}}AdvPC-\\ Latent\end{tabular} & \multicolumn{1}{c|}{\begin{tabular}[c]{@{}c@{}}BPDA-\\ PGD\end{tabular}} & \multicolumn{1}{c|}{\begin{tabular}[c]{@{}c@{}}EOT-\\ AutoAtt.\end{tabular}} & \multicolumn{1}{c|}{\begin{tabular}[c]{@{}c@{}}PGD-\\ Latent\end{tabular}} & \multicolumn{1}{c|}{\begin{tabular}[c]{@{}c@{}}AdvPC-\\ Latent\end{tabular}} & \multicolumn{1}{c|}{\begin{tabular}[c]{@{}c@{}}BPDA-\\ PGD\end{tabular}} & \multicolumn{1}{c|}{\begin{tabular}[c]{@{}c@{}}EOT-\\ AutoAtt.\end{tabular}} & \multicolumn{1}{c|}{\begin{tabular}[c]{@{}c@{}}PGD-\\ Latent\end{tabular}} & \begin{tabular}[c]{@{}c@{}}AdvPC-\\ Latent\end{tabular} \\ \hline
\multirow{6}{*}{{\begin{sideways}CurveNet\end{sideways}}} & \multirow{3}{*}{PointDP~\cite{sun2023critical}} & Clean Acc & 80.4 & 80.4 & 80.4 & \multicolumn{1}{c|}{80.4} & 80.4 & 80.4 & 80.4 & 80.4 & 45.0 & 45.0 & 45.0 & \multicolumn{1}{c|}{45.0} & 45.0 & 45.0 & 45.0 & 45.0 \\
 &  & Adv Acc & 65.0 & 67.7 & 65.4 & \multicolumn{1}{c|}{49.7} & 72.2 & 77.1 & 87.5 & 79.9 & 25.6 & 25.2 & 34.0 & \multicolumn{1}{c|}{22.9} & 38.9 & 43.4 & 55.5 & 47.7 \\
 &  & Mixed Acc & 72.7 & 74.1 & 72.9 & \multicolumn{1}{c|}{65.1} & 76.3 & 78.7 & 83.9 & 80.1 & 35.4 & 35.1 & 39.5 & \multicolumn{1}{c|}{34.0} & 42.0 & 44.2 & 50.3 & 46.3 \\ \cline{2-19} 
 & \multirow{3}{*}{PointDP+PST(Ours)} & Clean Acc & \textbf{84.2} & \textbf{83.2} & \textbf{84.3} & \multicolumn{1}{c|}{\textbf{84.4}} & \textbf{85.0} & \textbf{84.2} & \textbf{84.3} & \textbf{85.0} & \textbf{58.2} & \textbf{58.6} & \textbf{57.6} & \multicolumn{1}{c|}{\textbf{58.0}} & \textbf{61.1} & \textbf{59.9} & \textbf{59.9} & \textbf{58.7} \\
 &  & Adv Acc & \textbf{78.1} & \textbf{76.9} & \textbf{85.1} & \multicolumn{1}{c|}{\textbf{63.6}} & \textbf{79.6} & \textbf{82.4} & \textbf{89.4} & \textbf{84.2} & \textbf{49.0} & \textbf{44.9} & \textbf{53.7} & \multicolumn{1}{c|}{\textbf{37.9}} & \textbf{57.7} & \textbf{59.4} & \textbf{65.6} & \textbf{59.6} \\
 &  & Mixed Acc & \textbf{81.1} & \textbf{80.0} & \textbf{84.7} & \multicolumn{1}{c|}{\textbf{74.0}} & \textbf{82.3} & \textbf{83.3} & \textbf{86.8} & \textbf{84.6} & \textbf{53.6} & \textbf{51.8} & \textbf{55.6} & \multicolumn{1}{c|}{\textbf{48.0}} & \textbf{59.4} & \textbf{59.6} & \textbf{62.7} & \textbf{59.1} \\ \hline \hline
\multirow{6}{*}{{\begin{sideways}PointNet++\end{sideways}}} & \multirow{3}{*}{PointDP~\cite{sun2023critical}} & Clean Acc & 78.0 & 78.0 & 78.0 & \multicolumn{1}{c|}{78.0} & 78.0 & 78.0 & 78.0 & 78.0 & 47.4 & 47.4 & 47.4 & \multicolumn{1}{c|}{47.4} & 47.4 & 47.4 & 47.4 & 47.4 \\
 &  & Adv Acc & 64.5 & 66.1 & 45.7 & \multicolumn{1}{c|}{38.9} & 68.8 & 74.5 & 82.7 & 77.4 & 17.6 & 19.7 & 29.7 & \multicolumn{1}{c|}{16.5} & 28.1 & 44.9 & 50.6 & 47.5 \\
 &  & Mixed Acc & 71.3 & 72.1 & 61.9 & \multicolumn{1}{c|}{58.5} & 73.4 & 76.3 & 80.3 & 77.7 & 32.5 & 33.6 & 38.6 & \multicolumn{1}{c|}{31.9} & 37.8 & 46.1 & 49.0 & 47.5 \\ \cline{2-19} 
 & \multirow{3}{*}{PointDP+PST(Ours)} & Clean Acc & \textbf{82.7} & \textbf{83.9} & \textbf{82.9} & \multicolumn{1}{c|}{\textbf{82.4}} & \textbf{83.3} & \textbf{83.0} & \textbf{83.0} & \textbf{83.2} & \textbf{57.0} & \textbf{56.5} & \textbf{56.9} & \multicolumn{1}{c|}{\textbf{56.0}} & \textbf{56.6} & \textbf{57.2} & \textbf{57.8} & \textbf{56.0} \\
 &  & Adv Acc & \textbf{80.4} & \textbf{78.8} & \textbf{78.7} & \multicolumn{1}{c|}{\textbf{64.2}} & \textbf{79.2} & \textbf{81.7} & \textbf{89.0} & \textbf{84.0} & \textbf{47.2} & \textbf{45.4} & \textbf{54.3} & \multicolumn{1}{c|}{\textbf{39.2}} & \textbf{49.7} & \textbf{57.4} & \textbf{63.8} & \textbf{59.0} \\
 &  & Mixed Acc & \textbf{81.5} & \textbf{81.4} & \textbf{80.8} & \multicolumn{1}{c|}{\textbf{73.3}} & \textbf{81.3} & \textbf{82.3} & \textbf{86.0} & \textbf{83.6} & \textbf{52.1} & \textbf{51.0} & \textbf{55.6} & \multicolumn{1}{c|}{\textbf{47.6}} & \textbf{53.1} & \textbf{57.3} & \textbf{60.8} & \textbf{57.5} \\ \hline \hline
\multirow{6}{*}{{\begin{sideways}PCT\end{sideways}}} & \multirow{3}{*}{PointDP~\cite{sun2023critical}} & Clean Acc & 81.2 & 81.2 & 81.2 & \multicolumn{1}{c|}{81.2} & 81.2 & 81.2 & 81.2 & 81.2 & 47.5 & 47.5 & 47.5 & \multicolumn{1}{c|}{47.5} & 47.5 & 47.5 & 47.5 & 47.5 \\
 &  & Adv Acc & 69.1 & 70.5 & 68.0 & \multicolumn{1}{c|}{50.3} & 72.0 & 77.5 & \textbf{88.3} & 80.6 & 28.8 & 25.2 & 36.1 & \multicolumn{1}{c|}{22.9} & 38.2 & 46.3 & 56.5 & 49.7 \\
 &  & Mixed Acc & 75.2 & 75.8 & 74.6 & \multicolumn{1}{c|}{65.8} & 76.6 & 79.3 & 84.7 & 80.9 & 38.2 & 36.4 & 41.8 & \multicolumn{1}{c|}{35.2} & 42.9 & 46.9 & 52.1 & 48.6 \\ \cline{2-19} 
 & \multirow{3}{*}{PointDP+PST(Ours)} & Clean Acc & \textbf{82.6} & \textbf{84.0} & \textbf{84.3} & \multicolumn{1}{c|}{\textbf{83.8}} & \textbf{83.1} & \textbf{82.2} & \textbf{84.6} & \textbf{81.3} & \textbf{55.4} & \textbf{53.2} & \textbf{51.0} & \multicolumn{1}{c|}{\textbf{53.5}} & \textbf{55.1} & \textbf{56.9} & \textbf{57.0} & \textbf{56.7} \\
 &  & Adv Acc & \textbf{76.0} & \textbf{76.6} & \textbf{81.9} & \multicolumn{1}{c|}{\textbf{59.8}} & \textbf{77.0} & \textbf{79.6} & {87.7} & \textbf{81.2} & \textbf{46.9} & \textbf{42.4} & \textbf{44.3} & \multicolumn{1}{c|}{\textbf{39.1}} & \textbf{50.9} & \textbf{55.3} & \textbf{63.7} & \textbf{58.8} \\
 &  & Mixed Acc & \textbf{79.3} & \textbf{80.3} & \textbf{83.1} & \multicolumn{1}{c|}{\textbf{71.8}} & \textbf{80.0} & \textbf{81.0} & \textbf{86.2} & \textbf{81.3} & \textbf{51.2} & \textbf{47.8} & \textbf{47.6} & \multicolumn{1}{c|}{\textbf{46.3}} & \textbf{53.0} & \textbf{56.1} & \textbf{60.3} & \textbf{57.7} \\ \hline
\end{tabular}

  }  \label{tab:adaptive_single_attack}%
\end{table*}

\begin{table*}[!tbp]
  \centering
  \caption{Evaluation of robustness against adaptive attacks on ModelNet40 and ScanObjectNN datasets under Mixed Attack protocol.}
   \resizebox{1\linewidth}{!}{
\setlength{\tabcolsep}{4pt}
\begin{tabular}{clccccccccc|cccccccc}
\hline
\multirow{2}{*}{} & \multirow{2}{*}{} & \multirow{2}{*}{} & \multicolumn{8}{c|}{ModelNet40} & \multicolumn{8}{c}{ScanObjectNN} \\ \cline{4-19} 
 &  &  & \multicolumn{4}{c|}{\( l_\infty \)} & \multicolumn{4}{c|}{\( l_2 \)} & \multicolumn{4}{c|}{\( l_\infty \)} & \multicolumn{4}{c}{\( l_2 \)} \\ \cline{4-19} 
 &  &  & \multicolumn{1}{c|}{\begin{tabular}[c]{@{}c@{}}BPDA-\\ PGD\end{tabular}} & \multicolumn{1}{c|}{\begin{tabular}[c]{@{}c@{}}EOT-\\ AutoAtt.\end{tabular}} & \multicolumn{1}{c|}{\begin{tabular}[c]{@{}c@{}}PGD-\\ Latent\end{tabular}} & \multicolumn{1}{c|}{\begin{tabular}[c]{@{}c@{}}AdvPC-\\ Latent\end{tabular}} & \multicolumn{1}{c|}{\begin{tabular}[c]{@{}c@{}}BPDA-\\ PGD\end{tabular}} & \multicolumn{1}{c|}{\begin{tabular}[c]{@{}c@{}}EOT-\\ AutoAtt.\end{tabular}} & \multicolumn{1}{c|}{\begin{tabular}[c]{@{}c@{}}PGD-\\ Latent\end{tabular}} & \begin{tabular}[c]{@{}c@{}}AdvPC-\\ Latent\end{tabular} & \multicolumn{1}{c|}{\begin{tabular}[c]{@{}c@{}}BPDA-\\ PGD\end{tabular}} & \multicolumn{1}{c|}{\begin{tabular}[c]{@{}c@{}}EOT-\\ AutoAtt.\end{tabular}} & \multicolumn{1}{c|}{\begin{tabular}[c]{@{}c@{}}PGD-\\ Latent\end{tabular}} & \multicolumn{1}{c|}{\begin{tabular}[c]{@{}c@{}}AdvPC-\\ Latent\end{tabular}} & \multicolumn{1}{c|}{\begin{tabular}[c]{@{}c@{}}BPDA-\\ PGD\end{tabular}} & \multicolumn{1}{c|}{\begin{tabular}[c]{@{}c@{}}EOT-\\ AutoAtt.\end{tabular}} & \multicolumn{1}{c|}{\begin{tabular}[c]{@{}c@{}}PGD-\\ Latent\end{tabular}} & \begin{tabular}[c]{@{}c@{}}AdvPC-\\ Latent\end{tabular} \\ \hline
\multirow{6}{*}{{\begin{sideways}CurveNet\end{sideways}}} & \multirow{3}{*}{PointDP~\cite{sun2023critical}} & Clean Acc & \multicolumn{4}{c|}{72.3} & \multicolumn{4}{c|}{72.3} & \multicolumn{4}{c|}{47.9} & \multicolumn{4}{c}{47.9} \\
 &  & Adv Acc & 63.4 & 64.9 & 57.4 & \multicolumn{1}{c|}{64.5} & 67.9 & \textbf{74.4} & 92.0 & 93.1 & 25.0 & 28.6 & 34.2 & \multicolumn{1}{c|}{26.3} & 31.2 & 50.6 & 60.9 & 47.7 \\
 &  & Mixed Acc & \multicolumn{4}{c|}{64.6} & \multicolumn{4}{c|}{79.9} & \multicolumn{4}{c|}{32.4} & \multicolumn{4}{c}{47.7} \\ \cline{2-19} 
 & \multirow{3}{*}{PointDP+PST(Ours)} & Clean Acc & \multicolumn{4}{c|}{\textbf{81.3}} & \multicolumn{4}{c|}{\textbf{80.4}} & \multicolumn{4}{c|}{\textbf{50.2}} & \multicolumn{4}{c}{\textbf{59.4}} \\
 &  & Adv Acc & \textbf{76.5} & \textbf{68.2} & \textbf{85.8} & \multicolumn{1}{c|}{\textbf{84.0}} & \textbf{79.1} & {73.6} & \textbf{92.9} & \textbf{94.3} & \textbf{36.8} & \textbf{50.5} & \textbf{48.8} & \multicolumn{1}{c|}{\textbf{37.5}} & \textbf{43.8} & \textbf{59.7} & \textbf{66.7} & \textbf{58.8} \\
 &  & Mixed Acc & \multicolumn{4}{c|}{\textbf{79.1}} & \multicolumn{4}{c|}{\textbf{84.1}} & \multicolumn{4}{c|}{\textbf{44.8}} & \multicolumn{4}{c}{\textbf{57.7}} \\ \hline \hline
\multirow{6}{*}{{\begin{sideways}PointNet++\end{sideways}}} & \multirow{3}{*}{PointDP~\cite{sun2023critical}} & Clean Acc & \multicolumn{4}{c|}{70.3} & \multicolumn{4}{c|}{70.3} & \multicolumn{4}{c|}{42.8} & \multicolumn{4}{c}{42.8} \\
 &  & Adv Acc & 64.7 & 59.6 & 39.9 & \multicolumn{1}{c|}{43.8} & 68.1 & 69.7 & 87.0 & 90.8 & 21.1 & 27.6 & 30.0 & \multicolumn{1}{c|}{30.4} & 29.5 & 49.6 & 56.9 & 51.0 \\
 &  & Mixed Acc & \multicolumn{4}{c|}{55.7} & \multicolumn{4}{c|}{77.2} & \multicolumn{4}{c|}{30.4} & \multicolumn{4}{c}{46.0} \\ \cline{2-19} 
 & \multirow{3}{*}{PointDP+PST(Ours)} & Clean Acc & \multicolumn{4}{c|}{\textbf{75.8}} & \multicolumn{4}{c|}{\textbf{76.0}} & \multicolumn{4}{c|}{\textbf{59.2}} & \multicolumn{4}{c}{\textbf{61.1}} \\
 &  & Adv Acc & \textbf{77.5} & \textbf{69.4} & \textbf{80.5} & \multicolumn{1}{c|}{\textbf{81.0}} & \textbf{77.7} & \textbf{75.5} & \textbf{92.3} & \textbf{94.1} & \textbf{40.1} & \textbf{51.7} & \textbf{53.1} & \multicolumn{1}{c|}{\textbf{41.7}} & \textbf{46.0} & \textbf{64.4} & \textbf{71.9} & \textbf{55.9} \\
 &  & Mixed Acc & \multicolumn{4}{c|}{\textbf{76.8}} & \multicolumn{4}{c|}{\textbf{83.1}} & \multicolumn{4}{c|}{\textbf{49.2}} & \multicolumn{4}{c}{\textbf{59.9}} \\ \hline \hline
\multirow{6}{*}{{\begin{sideways}PCT\end{sideways}}} & \multirow{3}{*}{PointDP~\cite{sun2023critical}} & Clean Acc & \multicolumn{4}{c|}{74.0} & \multicolumn{4}{c|}{74.0} & \multicolumn{4}{c|}{57.2} & \multicolumn{4}{c}{57.2} \\
 &  & Adv Acc & 62.8 & 69.5 & 69.6 & \multicolumn{1}{c|}{62.8} & 63.1 & \textbf{74.8} & 92.4 & \textbf{95.1} & 13.8 & 35.9 & 24.5 & \multicolumn{1}{c|}{13.8} & 21.7 & 44.4 & 32.0 & 52.6 \\
 &  & Mixed Acc & \multicolumn{4}{c|}{67.7} & \multicolumn{4}{c|}{79.9} & \multicolumn{4}{c|}{29.1} & \multicolumn{4}{c}{41.6} \\ \cline{2-19} 
 & \multirow{3}{*}{PointDP+PST(Ours)} & Clean Acc & \multicolumn{4}{c|}{\textbf{77.8}} & \multicolumn{4}{c|}{\textbf{78.6}} & \multicolumn{4}{c|}{\textbf{69.6}} & \multicolumn{4}{c}{\textbf{68.2}} \\
 &  & Adv Acc & \textbf{67.3} & \textbf{71.4} & \textbf{87.8} & \multicolumn{1}{c|}{\textbf{80.9}} & \textbf{70.4} & {74.7} & \textbf{92.5} & {92.9} & \textbf{41.7} & \textbf{47.1} & \textbf{55.2} & \multicolumn{1}{c|}{\textbf{31.4}} & \textbf{44.4} & \textbf{61.8} & \textbf{67.1} & \textbf{53.0} \\
 &  & Mixed Acc & \multicolumn{4}{c|}{\textbf{77.1}} & \multicolumn{4}{c|}{\textbf{81.8}} & \multicolumn{4}{c|}{\textbf{49.0}} & \multicolumn{4}{c}{\textbf{58.9}} \\ \hline
\end{tabular}

  }  \label{tab:adaptive_mix_attack}%
\end{table*}

\noindent\textbf{Strong Adaptive Attacks}:
We further evaluate on strong adaptive attacks with results on ModelNet40 and ScanObjectNN datasets with Single and Mixed Attacks. Following the protocol in PointDP~\cite{sun2023critical}, we assume that attackers have access to the classification outputs of samples after passing through the diffusion model. The adversarial perturbation is implemented at the latent features of PointDP. The results are reported in Tab.~\ref{tab:adaptive_single_attack} and Tab.~\ref{tab:adaptive_mix_attack} for single attack and mixed attack respectively. We observe from the results that our proposed method consistently improves both clean accuracy and adversarial accuracy with a significant margin. The improvement on ScanObjectNN is more obvious with 10-20$\%$ increase in mixed accuracy under Mixed Attack protocol. The results indicate the effectiveness of self-training procedure without relying on a safely purified sample.

\begin{figure}[!tbp]
    \centering
    \includegraphics[width=0.99\linewidth]{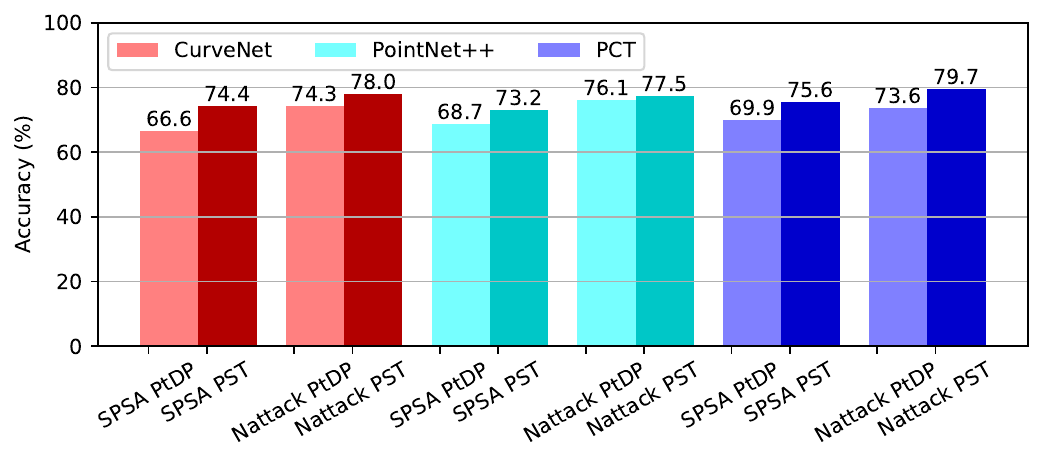}
    \caption{Adversarial robustness of black-box attacks on ModelNet40 under $\ell_\infty$ 
constraint. PtDP and PST refer to PointDiffusion and Purified Self-Training~(Ours) respectively.}
    \label{fig:black_box}
\end{figure}

\noindent\textbf{Black-Box Attacks}:
We evaluate additional black-box attacks to verify whether the success of self-training is due to hiding the dynamic model to adversarial attacks. Specifically, we evaluate two strong black-box attacks, SPSA and Nattack on PointNet\texttt{++} and PCT backbones. We followed the approach of PointDP and utilized a subset of 512 samples to assess the model's robustness. We observe from Fig.~\ref{fig:black_box} that our self-training consistently improves from PointDP purification alone, suggesting the effectiveness of self-training is not due to hiding the dynamic model from the adversarial attacks.

\begin{table*}[!tbp]
  \centering
  \caption{Ablation study on ModelNet40 dataset with PCT and PointNet++
backbones.}
   \resizebox{0.85\linewidth}{!}{
\setlength{\tabcolsep}{3pt} 
\begin{tabular}{ccccccccc|cccccc}
\hline
\multicolumn{3}{l}{\multirow{2}{*}{}} & \multicolumn{6}{c|}{PCT} & \multicolumn{6}{c}{PointNet++} \\ \cline{4-15} 
\multicolumn{3}{l}{} & \multicolumn{3}{c|}{White-box Attack} & \multicolumn{3}{c|}{Adaptive Attack} & \multicolumn{3}{c|}{White-box Attack} & \multicolumn{3}{c}{Adaptive Attack} \\ \hline
\multicolumn{1}{c|}{Self-Training} & \multicolumn{1}{c|}{\begin{tabular}[c]{@{}c@{}}Feature \\ Distribution \\ Alignment\end{tabular}} & \multicolumn{1}{c|}{\begin{tabular}[c]{@{}c@{}}Point\\ Diffusion\end{tabular}} & \multicolumn{1}{c|}{\begin{tabular}[c]{@{}c@{}}Clean\\ Acc.\end{tabular}} & \multicolumn{1}{c|}{\begin{tabular}[c]{@{}c@{}}Avg.\\ Adv. Acc.\end{tabular}} & \multicolumn{1}{c|}{\begin{tabular}[c]{@{}c@{}}Mixed\\ Acc.\end{tabular}} & \multicolumn{1}{c|}{\begin{tabular}[c]{@{}c@{}}Clean\\ Acc.\end{tabular}} & \multicolumn{1}{c|}{\begin{tabular}[c]{@{}c@{}}Avg.\\ Adv. Acc.\end{tabular}} & \begin{tabular}[c]{@{}c@{}}Mixed\\ Acc.\end{tabular} & \multicolumn{1}{c|}{\begin{tabular}[c]{@{}c@{}}Clean\\ Acc.\end{tabular}} & \multicolumn{1}{c|}{\begin{tabular}[c]{@{}c@{}}Avg.\\ Adv. Acc.\end{tabular}} & \multicolumn{1}{c|}{\begin{tabular}[c]{@{}c@{}}Mixed\\ Acc.\end{tabular}} & \multicolumn{1}{c|}{\begin{tabular}[c]{@{}c@{}}Clean\\ Acc.\end{tabular}} & \multicolumn{1}{c|}{\begin{tabular}[c]{@{}c@{}}Avg.\\ Adv. Acc.\end{tabular}} & \begin{tabular}[c]{@{}c@{}}Mixed\\ Acc.\end{tabular} \\ \hline
\multicolumn{1}{c|}{-} & \multicolumn{1}{c|}{-} & \multicolumn{1}{c|}{-} & \multicolumn{1}{c|}{\textbf{89.3}} & \multicolumn{1}{c|}{30.6} & \multicolumn{1}{c|}{42.4} & \multicolumn{1}{c|}{89.3} & \multicolumn{1}{c|}{53.3} & 60.5 & \multicolumn{1}{c|}{89.3} & \multicolumn{1}{c|}{30.5} & \multicolumn{1}{c|}{42.3} & \multicolumn{1}{c|}{89.3} & \multicolumn{1}{c|}{41.3} & 51.0 \\ \hline
\multicolumn{1}{c|}{\begin{tabular}[c]{@{}c@{}}Fixed\\ Threshold\end{tabular}} & \multicolumn{1}{c|}{-} & \multicolumn{1}{c|}{-} & \multicolumn{1}{c|}{88.9} & \multicolumn{1}{c|}{30.7} & \multicolumn{1}{c|}{42.4} & \multicolumn{1}{c|}{87.9} & \multicolumn{1}{c|}{56.1} & 62.5 & \multicolumn{1}{c|}{88.9} & \multicolumn{1}{c|}{33.6} & \multicolumn{1}{c|}{44.7} & \multicolumn{1}{c|}{\textbf{90.5}} & \multicolumn{1}{c|}{59.8} & 66.0 \\ \hline
\multicolumn{1}{c|}{\begin{tabular}[c]{@{}c@{}}Adaptive\\ Threshold\end{tabular}} & \multicolumn{1}{c|}{-} & \multicolumn{1}{c|}{-} & \multicolumn{1}{c|}{87.9} & \multicolumn{1}{c|}{33.8} & \multicolumn{1}{c|}{44.6} & \multicolumn{1}{c|}{85.3} & \multicolumn{1}{c|}{59.8} & 64.9 & \multicolumn{1}{c|}{89.1} & \multicolumn{1}{c|}{34.3} & \multicolumn{1}{c|}{45.3} & \multicolumn{1}{c|}{89.7} & \multicolumn{1}{c|}{63.4} & 68.7 \\ \hline
\multicolumn{1}{c|}{\begin{tabular}[c]{@{}c@{}}Adaptive\\ Threshold\end{tabular}} & \multicolumn{1}{c|}{$\checkmark$} & \multicolumn{1}{c|}{-} & \multicolumn{1}{c|}{86.6} & \multicolumn{1}{c|}{47.6} & \multicolumn{1}{c|}{55.5} & \multicolumn{1}{c|}{\textbf{89.7}} & \multicolumn{1}{c|}{68.5} & 72.8 & \multicolumn{1}{c|}{\textbf{89.3}} & \multicolumn{1}{c|}{40.7} & \multicolumn{1}{c|}{50.5} & \multicolumn{1}{c|}{89.3} & \multicolumn{1}{c|}{69.8} & 73.7 \\ \hline
\multicolumn{1}{c|}{-} & \multicolumn{1}{c|}{-} & \multicolumn{1}{c|}{PointDP} & \multicolumn{1}{c|}{73.9} & \multicolumn{1}{c|}{74.7} & \multicolumn{1}{c|}{74.5} & \multicolumn{1}{c|}{74.0} & \multicolumn{1}{c|}{66.2} & 67.7 & \multicolumn{1}{c|}{70.3} & \multicolumn{1}{c|}{71.2} & \multicolumn{1}{c|}{71.0} & \multicolumn{1}{c|}{70.3} & \multicolumn{1}{c|}{52.0} & 55.7 \\ \hline
\multicolumn{1}{c|}{\begin{tabular}[c]{@{}c@{}}Adaptive\\ Threshold\end{tabular}} & \multicolumn{1}{c|}{$\checkmark$} & \multicolumn{1}{c|}{PointDP} & \multicolumn{1}{c|}{77.6} & \multicolumn{1}{c|}{\textbf{80.1}} & \multicolumn{1}{c|}{\textbf{79.6}} & \multicolumn{1}{c|}{77.8} & \multicolumn{1}{c|}{\textbf{76.9}} & \textbf{77.1} & \multicolumn{1}{c|}{75.0} & \multicolumn{1}{c|}{\textbf{82.9}} & \multicolumn{1}{c|}{\textbf{81.3}} & \multicolumn{1}{c|}{75.8} & \multicolumn{1}{c|}{\textbf{77.1}} & \textbf{76.8} \\ \hline
\end{tabular}
  }  \label{tab:ablation}%
\end{table*}

\noindent\textbf{Transferability of Attacks}: 
We further investigate whether the adversarial attack could transfer across different backbones and compare Purified Self-Training with PointDP. Specifically, we evaluate $l_\infty$ attacks under mixed attack protocol. Both white-box and adaptive attacks are evaluated on ModelNet40 and ScanObjectNN datasets. We make the following observations from the results in Tab.~\ref{tab:transferability}. First, our method significantly outperforms PointDP with all backbone pairs, suggesting the effectiveness of test-time training for improving adversarial robustness. Moreover, we observe the adversarial attacks are transferrable across backbones, suggesting the adversarial attacks are real threats to the robustness of 3D point cloud models.

\begin{table}[htbp]
  \centering
  \caption{Evaluation of transfer attacks on ModelNet40 and ScanObjNN with whitebox attacks. Each row indicate the backbone used for testing and each column indicate the backbone used for generating adversarial attack. The mixed accuracies for PointDP and PointDP+PST~(ours) are delimitered by /.}
  \setlength{\tabcolsep}{2pt}
   \resizebox{0.99\linewidth}{!}{
    \begin{tabular}{lccc|ccc}
    \toprule
          & \multicolumn{3}{c|}{ModelNet40} & \multicolumn{3}{c}{ScanObjNN} \\
\cmidrule{2-7}          & \multicolumn{1}{c}{CurveNet} & \multicolumn{1}{c}{PointNet++} & \multicolumn{1}{c|}{PCT} & \multicolumn{1}{c}{CurveNet} & \multicolumn{1}{c}{PointNet++} & \multicolumn{1}{c}{PCT} \\
    \midrule
    CurveNet & 73.2/81.6 & 75.8/83.5 & 75.0/81.7 & 31.6/49.7 & 34.0/51.9 & 32.4/49.3 \\
    PointNet++ & 69.8/80.3 & 71.0/81.3 & 69.8/80.2 & 30.2/50.4 & 38.1/49.6 & 32.3/44.3 \\
    PCT   & 77.8/79.2 & 77.8/79.5 & 74.5/79.6 & 35.8/45.4 & 38.1/49.6 & 32.3/44.3 \\
    \bottomrule
    \end{tabular}%
    }
  \label{tab:transferability}%
\end{table}%

\subsection{{Qualitative Results}}

\begin{figure*}
    \centering
    \includegraphics[width=1.05\linewidth]{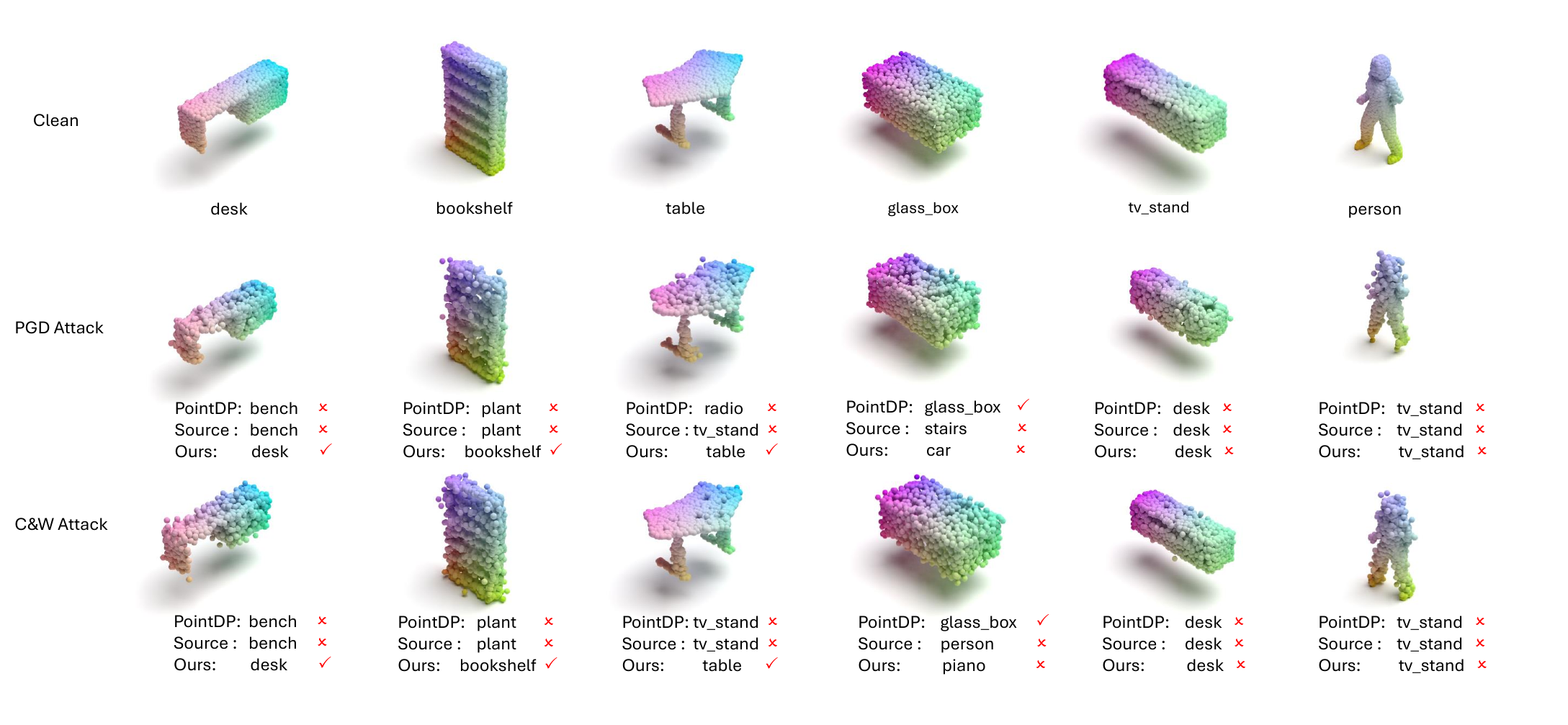}
    \caption{Qualitative results of adversarial attacks on different objects and defense results. Each row indicates one type of attack or w/o attack (Clean). }
    \label{fig:Qualitative Results}
\end{figure*}

We include qualitative results for the adversarial attacks and the prediction results by competing methods in Fig.~\ref{fig:Qualitative Results}. We make the following observations from the results. Despite all methods making correct predictions on clean samples, the attacked samples present challenges to all methods. In general our proposed PST achieves better predictions for attacked samples. The mistakes are mainly made on easily confusing classes, e.g. classifying ``tv\_stand'' to ``desk''.

\subsection{Ablation Study}

We investigate the effectiveness of proposed self-training procedure by ablating the following components including self-training, feature distribution alignment and purification. We conduct ablation study with white-box and adaptive attacks. For white-box attack, we report the average accuracy over four types of attacks~(C\&W, PGD, AdvPC \& SI\_Adv). For adaptive attack, we report the average accuracy over four types of attacks~(BPDA-PGD \cite{athalye2018obfuscated}, EOT-AutoAttack~(EOT-AutoAtt.), PGD-Latent \& AdvPC-Latent). We present the ablation study in Tab.~\ref{tab:ablation} and make the following observations. First, we observe that self-training with either fixed threshold or adaptive threshold makes only marginal improvement. This suggests that self-training alone could be prone to incorrect pseudo labels, thus limiting the adversarial robustness. When feature distribution alignment is further included, we observe a significant improvement, e.g. $44.6\%$ to $55.5\%$ on white-box attack with PCT. Alternatively, if point diffusion is applied alone, which is equivalent to PointDP~\cite{sun2023critical}, we observe relatively improved robustness against white-box attack. Finally, when our self-training is combined with PointDP purification, we witness the most impressive results. The improvement from PointDP alone is very significant on the stronger adaptive attacks, suggesting updating model on-the-fly through self-training is a promising approach to be used in conjunction with adversarial purificationeon.
\vspace{-0.3cm}

\subsection{Test-Time Performance Analysis}

\begin{figure}[!tb] 
\centering 
\hspace{-0.3cm}
\begin{subfigure}{0.49\linewidth}  
\centering  
\includegraphics[width=1.12\linewidth]{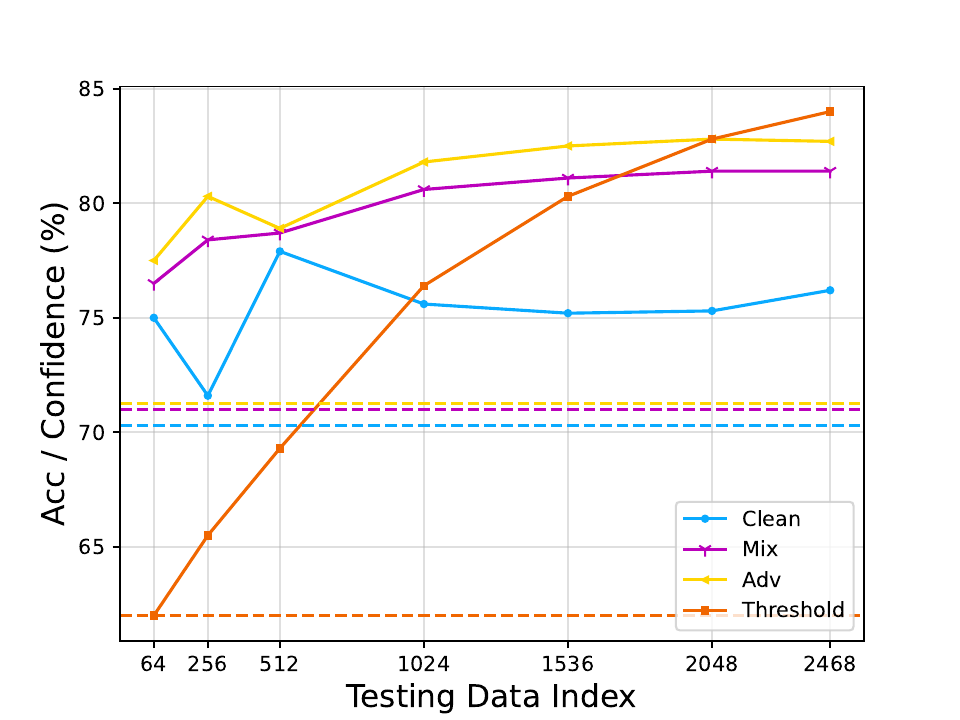}  \caption{ModelNet40} 
\end{subfigure} \hspace{-0.1cm}
\begin{subfigure}{0.49\linewidth}  
\centering  
\includegraphics[width=1.12\linewidth]{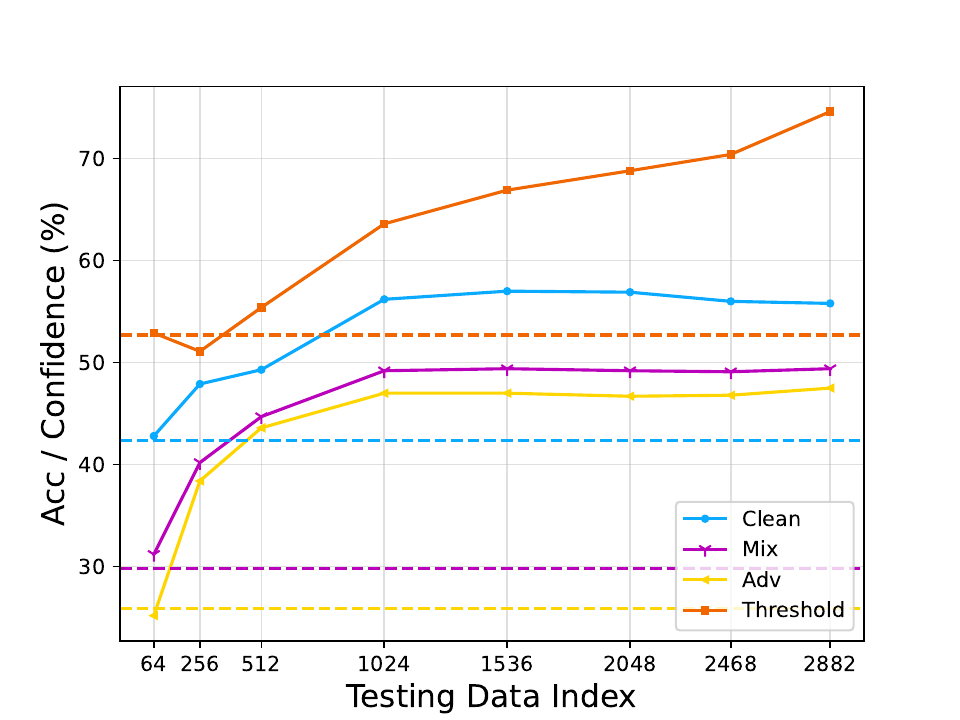}  \caption{ScanObjectNN} 
\end{subfigure} 
\caption{The evolution of model accuracy across testing data streams with PointNet\texttt{++} backbone. Dashed and solid lines correspond to PointDP and our method, respectively. } 
\label{figtest}
\vspace{-0.5cm}
\end{figure}

We visualize the evolution of our model's accuracy as samples arrived in the stream, as shown in Fig.~\ref{figtest}. It is evident that, compared to the PointDP baseline, our model achieves a substantial improvement in accuracy from the beginning (index 64). As more samples arrive, the accuracy of our model gradually increases. However, the accuracy for clean data plateaus, which is likely attributed to the semantic losses in the reconstruction process, indicating that the model's fitting capacity may have saturated. We also present the evolution of global confidence~(Threshold). The Threshold experiences a monotonic increase, suggesting the model becomes more confident upon self-training on testing data and an adaptive threshold is necessary to accommodate to the increase of model confidence.

\section{Conclusion}
\vspace{-0.3cm}
This work introduces a novel approach for enhancing the robustness of 3D point cloud deep learning models against adversarial attacks during the inference stage. We propose an innovative inference stage adaptation method, employing a purified self-training procedure on a diffusion-based model. This dynamic adaptation involves generating high-confidence predictions as pseudo-labels and implementing distribution alignment and adaptive thresholding to improve robustness, as well as combining test-time purification with test-time training. To align with the setting of real-world applications, we also present a novel evaluation protocol simulating realistic adversarial attacks in a streaming fashion, demonstrating the adaptability of our method to evolving attack scenarios.

\noindent\textbf{Acknowledgement}: This work is supported by A*STAR under Project
M23L7b0021 and Sichuan Science and Technology Program (Project No.: 2023NSFSC1421).

\bibliographystyle{elsarticle-harv} 
\bibliography{ref}

\end{document}